\documentclass[review]{elsarticle}

\usepackage[margin=1in]{geometry}
\usepackage{float}
\usepackage{amssymb}
\usepackage{booktabs}
\usepackage{amsmath}
\usepackage{makecell}
\usepackage{mathtools}
\usepackage{subcaption}
\usepackage{lineno}
\usepackage[table]{xcolor}
\usepackage{multirow}
\usepackage{siunitx}
\usepackage{caption}
\usepackage{xparse}
\usepackage{adjustbox}
\usepackage{longtable}
\usepackage{tikz}
\usetikzlibrary{positioning, shapes.geometric, calc}

\ExplSyntaxOn
\NewDocumentCommand{\colorcell}{mm}{
    \cellcolor{ green! \fp_eval:n { (#1)^3 * 100 } !white }
    \num{ #2 }
}
\ExplSyntaxOff

\newcommand{\HPGe}{HPGe}
\newcommand{\LaBr}{LaBr$_3$}
\newcommand{\NaI}{NaI(Tl)}
\newcommand{\gadras}{\texttt{GADRAS}}
\newcommand{\geant}{\texttt{Geant4}}

\makeatletter
\def\ps@pprintTitle{
 \let\@oddhead\@empty
 \let\@evenhead\@empty
 \def\@oddfoot{\centerline{\thepage}}
 \let\@evenfoot\@oddfoot}
\makeatother

\begin{document}

\begin{frontmatter}

\title{Sim-to-real supervised domain adaptation for radioisotope identification}

\author[PNNLaddress]{Peter Lalor\corref{corauthor}}
\author[PNNLaddress,UWaddress]{Henry Adams}
\author[PNNLaddress]{Alex Hagen}

\cortext[corauthor]{Corresponding author
\\\hspace*{13pt} Email address: peter.lalor@pnnl.gov
\\\hspace*{13pt} Telephone: (925) 453-1876}
\address[PNNLaddress]{Pacific Northwest National Laboratory, Richland, WA 99352 USA}
\address[UWaddress]{University of Washington, Seattle, WA 98195 USA}

\begin{abstract}
Machine learning has the potential to improve the speed and reliability of radioisotope identification using gamma spectroscopy. However, meticulously labeling an experimental dataset for training is often prohibitively expensive, while training models purely on synthetic data is risky due to the domain gap between simulated and experimental measurements. In this research, we demonstrate that supervised domain adaptation can substantially improve the performance of radioisotope identification models by transferring knowledge between synthetic and experimental data domains. We consider two domain adaptation scenarios: (1) a simulation-to-simulation adaptation, where we perform multi-label proportion estimation using simulated high-purity germanium detectors, and (2) a simulation-to-experimental adaptation, where we perform multi-class, single-label classification using measured spectra from handheld lanthanum bromide (\LaBr) and sodium iodide (\NaI) detectors. We begin by pretraining a spectral classifier on synthetic data using a custom transformer-based neural network. After subsequent fine-tuning on just 64 labeled experimental spectra, we achieve a test accuracy of 96\% in the sim-to-real scenario with a {\LaBr} detector, far surpassing a synthetic-only baseline model (75\%) and a model trained from scratch (80\%) on the same 64 spectra. Furthermore, we demonstrate that domain-adapted models learn more human-interpretable features than experiment-only baseline models. Overall, our results highlight the potential for supervised domain adaptation techniques to bridge the sim-to-real gap in radioisotope identification, enabling the development of accurate and explainable classifiers even in real-world scenarios where access to experimental data is limited.
\end{abstract}
\begin{keyword}
Domain Adaptation, Transformer, Neural Network, Gamma Spectroscopy
\end{keyword}
\end{frontmatter}

\begin{sloppypar}

\section{Introduction}
\label{Introduction}

Spectroscopy and spectrometry are widely used characterization methods, and depending on the spectrum type, can illuminate details about a material's crystallographic, molecular, elemental, or isotopic composition. The applications of such spectral characterization and analysis span fields including high-energy physics~\cite{nonakaPileupCorrectionsHigherorder2020}, remote sensing~\cite{gaoNovelUnsupervisedSegmentation2017}, biology~\cite{fliriBiologicalSpectraAnalysis2005}, and nuclear material analysis~\cite{Hagen2022}. While the motivations, physical mechanisms, and equipment used vary widely across spectral analysis, they are all unified by the processing of a ``spectrum''--the histogram of discrete counts of a quantum arriving at a detector versus a physical quantity (usually energy or wavelength) indicative of each quantum over time. In the present work, we focus on gamma spectroscopy, which measures a spectrum of the number of gamma photons depositing energy in a detector across discrete ranges of deposited energy~\cite{knoll2010radiation}. While we focus on specifics of gamma spectroscopy, the analysis of how to best enable spectral simulation-based inference is more general.

Many detector types have been developed for detecting gammas and their energy, each with unique resolution (a measurement of the ability to discriminate between small differences in gamma energy), efficiency (a measurement of the proportion of gammas incident on the detector that are detected), and calibration (a description of how the deposition energy is converted to an electrical signal and eventually measured). Traditional techniques for gamma spectral analysis search for peaks in the measured gamma energy spectrum that can then be mapped to specific decay energies of a radioisotope~\cite{sampson2007plutonium, genie2000, peakeasy}. These methods typically include numerous time-consuming preprocessing steps and require close monitoring by an expert spectroscopist~\cite{elliot2020peakmap, darweesh2019study, croce2021improved}. Furthermore, classical approaches struggle in the presence of thick shielding or when the spectrum contains many overlapping material signatures. Supervised machine learning is an excellent alternative candidate for spectral analysis due to its ability to learn near-optimal representations of data for a given task~\cite{Goodfellow2016}. Gamma spectroscopy, however, introduces some challenges to the adoption of supervised machine learning. First and foremost, supervised machine learning requires large labeled datasets for high performance, and so-called ``neural scaling laws'' have emerged showing enormous scale is required for difficult tasks (e.g., 38 million images per class for 99\% accurate image classification on natural images~\cite{hestnessDeepLearningScaling2017}). The collection of such large datasets of gamma spectroscopy data would be extremely costly and time-consuming. Further, the labeling of a gamma spectrum is ambiguous. At the spectrum level, if a complete spectrum is labeled according to a recorded set of experimenter-placed radioactive sources, these labels will be incomplete due to naturally occurring radioactive material (NORM) in the background and cosmic radiation sources; at the quanta level, it is presently impossible to label each detection event according to its origin source from among the nearby radioactive sources, NORM, and cosmic sources~\cite{fernandezImprovementNuclideDetection2025}.

This has led to a trend in the field of supervised learning for spectroscopic analysis where simulation-based inference is leveraged~\cite{fernandezImprovementNuclideDetection2025, cranmerFrontierSimulationbasedInference2020, archambault2023g4ares, pierson2022alpha}. Amongst simulation-based inference techniques, naive application to real data of a classifier trained on simulated data incurs a huge misclassification risk, since the ability of such models to generalize is inherently limited by the difference in the data distributions between those generated by simulation and those generated by real processes, hereafter the ``sim-to-real gap''. In this analysis, we consider the scenario where we have access to a data-rich source domain (simulations) and wish to transfer knowledge to perform classification in a data-scarce target domain (experimental measurements). In the present work, we characterize the effectiveness of this approach, leaving the investigation of other simulation-based inference approaches, such as the iterative approaches from Ref.~\cite{cranmerFrontierSimulationbasedInference2020} or unsupervised and semi-supervised approaches~\cite{tarvainen2018meanteachersbetterrole} for future studies. As such, we consider two primary research questions:

\begin{enumerate}
  \item Does pretraining followed by fine-tuning yield a statistically significant improvement in model performance over models trained (a) only on source-domain data and (b) only on target-domain data?
  \item Do different architectures produce statistical differences in the chosen evaluation metrics?
\end{enumerate}

Furthermore, we are interested in how the answer to these questions depends on (i) target domain dataset size, (ii) simulated versus experimental target domain, and (iii) detector type (\HPGe, \LaBr, \NaI).

\section{Related Work}
\label{Related Work}

Machine learning methods were first used for radioisotope identification by Olmos \emph{et al.}~\cite{Olmos1991, Olmos1992}, where the authors demonstrated that artificial neural networks (ANNs) can achieve high classification accuracy on experimental gamma spectra. More recently, Liang \emph{et al.} trained a convolutional neural network (CNN) using simulated mixed sources, achieving high accuracy on low-resolution spectra~\cite{Liang2019rapid}. Kamuda developed an open-source Python package \texttt{annsa}, capable of creating simulated gamma-ray training datasets and applying machine learning models to solve spectroscopic tasks~\cite{kamuda2019automated}. The author built upon this research in Ref.~\cite{kamuda2020comparison}, finding that CNNs had smaller variance in their output than multilayer perceptrons (MLPs).

Moore \emph{et al.} applied the CNN architecture to time-sequenced gamma spectra by analyzing a two-dimensional waterfall plot~\cite{Moore2019application}. The authors continued this research in Ref.~\cite{moore2020transfer}, where they achieved mixed results leveraging transfer learning techniques to combine simulated data with experimental measurements. Daniel \emph{et al.} also considered the task of out-of-distribution (OOD) testing, achieving ${>}90\%$ accuracy on real data after training a CNN on a fully synthetic database~\cite{daniel2020automatic}.

Bilton \emph{et al.} examined the use of ANNs for mobile detection using simulated {\NaI} detectors, finding that autoencoder-based spectral anomaly detection outperforms principal component analysis (PCA) and non-negative matrix factorization by up to $23\%$ and recurrent neural networks (RNNs) enhanced source identification accuracy by $17\%$ over static ANNs~\cite{bilton2021neural}. Ghawaly \emph{et al.} similarly trained a deep convolutional autoencoder, outperforming PCA-based methods at flagging anomalous gamma spectra in both real-world and synthetic urban environments~\cite{ghawaly2022characterization}.

Khatiwada \emph{et al.} surveyed different machine learning algorithms for emergency response applications. The authors found that models trained on synthetic datasets offered excellent performance on simulated test sets, and performance was comparable to conventional methods when applied to an experimental dataset~\cite{khatiwada2023machine}. Li \emph{et al.} introduced a more advanced transformer-based neural network for nuclide identification, achieving an improved recognition rate compared to classical architectures~\cite{Li2024}. The authors reshaped 1024-channel gamma spectra into $32 \times 32$ patches, and the z-score-normalized counts of each patch were used as the embedding vector for the subsequent attention layers.

Morrow \emph{et al.} developed an open-source Python package \texttt{PyRIID} for generating synthetic gamma spectra for training machine learning models~\cite{Morrow2021PyRIID}. Van Omen \emph{et al.} used \texttt{PyRIID} to synthesize short-lived fission products to train a model using a semi-supervised loss function, aiming to perform OOD multi-label proportion estimation~\cite{vanomen2024multilabel}. Barradas \emph{et al.} similarly used high-quality synthetic spectra to train a complex ANN, outperforming conventional peak-recognition methods when identifying up to ten radionuclides in multi-source {\LaBr} gamma spectra~\cite{Barradas2025}.

\section{Dataset Curation}
\label{Dataset Curation}

We consider six datasets, divided into three domain adaptation scenarios, summarized in Table~\ref{table:datasets}. In the first scenario (sim-to-sim adaptation), the task is multi-label proportion estimation on a simulated dataset of mixed \geant{} spectra using a {\HPGe} detector. In the second and third scenarios (sim-to-real adaptation), the task is multi-class, single-label classification on a dataset of experimental handheld spectra taken using {\LaBr} and {\NaI} detectors.

\begin{table}
\centering
\caption{Summary of the datasets used in this study.}
\begin{tabular}{@{}lllccc@{}}
\toprule
\textbf{Scenario} 
  & \textbf{Domain} 
  & \makecell{\textbf{Acquisition}\\[-0.5ex]\textbf{Mode}} 
  & \makecell{\textbf{Isotopic}\\[-0.5ex]\textbf{Composition}} 
  & \makecell{\textbf{Number of}\\[-0.5ex]\textbf{Isotopes}} 
  & \textbf{Size} \\
\midrule
\multirow{2}{*}{sim-to-sim (\HPGe)}
  & Source & \gadras{}     & Mixed  & 55 & $1.3\times10^6$ \\
  & Target & \geant{}     & Mixed  & 55 & $1.3\times10^6$ \\
\addlinespace
\multirow{2}{*}{sim-to-real (\LaBr)}
  & Source & \gadras{}     & Single & 32 & $1.4\times10^6$ \\
  & Target & Experiment & Single & 32 & 15,091        \\
\addlinespace
\multirow{2}{*}{sim-to-real (\NaI)}
  & Source & \gadras{}     & Single & 32 & $1.4\times10^6$ \\
  & Target & Experiment & Single & 32 & 10,440        \\
\bottomrule
\end{tabular}
\label{table:datasets}
\end{table}

\subsection{Sim-to-sim adaptation}

In the first scenario, we simulated a dataset of gamma spectra using \gadras{} (source domain) and aimed to perform isotopic identification on a spectral dataset simulated in \geant{} (target domain). \gadras{} (Gamma Detector Response and Analysis Software~\cite{horne2016gadras}) is a fast, deterministic, semi-empirical package for analyzing gamma-ray detector responses to user-specific sources. \geant{} (GEometry ANd Tracking~\cite{allison2016recent}) is a general-purpose, open-source Monte Carlo toolkit for simulating the passage of particles through matter, capable of modeling detector responses in complex 3D geometries.

In \geant{}, we simulated the gamma emission spectra of 55 individual radioisotopes by uniformly dissolving each isotope in a scintillation cocktail. This source configuration mirrors the experimental arrangement reported in Ref.~\cite{pierson2022alpha} and was selected to capture realistic self-attenuation and geometric effects. A high-purity germanium (\HPGe) detector was placed one meter from the source, and the energy of all incoming gamma rays was recorded and binned to create a gamma spectrum. We leveraged \texttt{G4ARES} (Geant4 Advanced Radio-Emission Simulation Framework~\cite{archambault2023g4ares}) to improve estimations of gamma cascade summing corrections. In post-processing, we applied a low-energy cutoff and performed Gaussian energy broadening over each spectrum~\cite{knoll2010radiation}. Furthermore, we introduced slight variations to energy calibration and detector resolution between different detector setups. Once these template spectra were simulated, we created mixed spectra by arbitrarily summing up to 14 individual spectra with random proportions. Our choice to include multiple isotopes per spectrum captures the common scenario in which several radioisotopes are simultaneously present in a material sample. We subsequently added background (cosmic, $^{40}$K, $^{226}$Ra, and $^{232}$Th) and Poisson noise. We synthesized a total of $10^6$ training spectra, along with separate holdout sets for validation and testing. To clarify, we did not use the entire dataset for model training; instead, we trained on data subsets of varying sizes to compare the performance of different models (with and without pretraining) as a function of dataset size.

We followed a similar sequence of steps for generating \gadras{} data, first simulating 55 template spectra using a 95\% efficiency {\HPGe} detector (relative to a 3$\times$3-inch {\NaI}). The spectra were then mixed, background-added, and Poisson resampled to produce a total of $10^6$ training spectra, performed using the \texttt{PyRIID} \texttt{SeedMixer} and \texttt{StaticSynthesizer} classes~\cite{Morrow2021PyRIID}. In both \gadras{} and \geant{}, all spectra consisted of 1024 uniform energy bins spanning from 0 to 3000~keV, with slight endpoint fluctuations due to variations in energy calibration.

\subsection{Sim-to-real adaptation}

In the experimental domain adaptation scenarios, we again used \gadras{} as our source domain, but sought to perform isotopic classification using experimental spectra. The experimental dataset was collected in 2018 at Pacific Northwest National Laboratory to characterize the performance of handheld {\LaBr} and {\NaI} detectors. Thirty-two different source isotopes were employed in standardized source-shield configurations (none, steel, lead, polyethylene). The data was parsed using \texttt{SpecUtils}, a library for opening and manipulating spectrum files~\cite{SpecUtils-github}, and subsequently preprocessed to remove spectra that were corrupted, empty, or missing metadata. Furthermore, all spectra were interpolated onto a uniform 1024-bin energy grid from 0 to 3000~keV. This resulted in a labeled dataset of 15,091 {\LaBr} spectra and 10,440 {\NaI} spectra, partitioned into training, validation, and test sets with a 70/15/15 split. As before, we only considered subsets of the training data at a time to benchmark performance as a function of training size.

The experimental datasets used in this study are independent of the sim-to-sim \geant{} simulations and reflect a different measurement scenario. Furthermore, we remark that the experimental datasets available for this study are relatively simple, and 100\% testing accuracy can be achieved while only training on $\approx 500$ training examples due to the nature of the data collection process (single-label, high signal-to-noise ratio). While a more complicated experimental dataset is desirable, we still expect the results of this research to apply broadly and refer to the sim-to-sim adaptation as a case study using complex, mixed sources.

The \gadras{} dataset was synthesized to replicate the experimental configuration by selecting the corresponding built‑in handheld {\LaBr}/{\NaI} detector model from the \gadras{} detector library. Spectra were simulated for each of the 32 radioisotopes over a range of source-to-detector distances, detector heights, energy calibrations, and shielding configurations. These spectra were then mixed with \gadras{}-simulated background (cosmic, $^{40}$K, $^{226}$Ra, and $^{232}$Th) to create seed templates using the \texttt{PyRIID} \texttt{SeedMixer} class. The mixing fractions were drawn from a Dirichlet distribution whose parameters were estimated by fitting measured background spectra as a non-negative combination of the four simulated component spectra. These mixed spectrum seed templates were then used to create a synthetic dataset using the \texttt{PyRIID} \texttt{StaticSynthesizer} class, varying over parameters such as background count rate, live time, and signal-to-noise ratio, with the values of each parameter approximated empirically. In total, we synthesized $10^6$ training spectra, $2 \times 10^5$ validation spectra, and $2 \times 10^5$ testing spectra in less than 15 minutes on a commodity laptop. This process was repeated for each detector type (\LaBr, \NaI).

\section{Method}
\label{Method}

\subsection{Model Architectures}
\label{architectures}

We implemented four machine learning architectures for this analysis. MLPs and CNNs are commonly used in the gamma-spectroscopy literature and serve as simple, effective benchmarks. We also consider attention-based models (first introduced for neural machine translation~\cite{Vaswani2017}) and include a prior transformer-based neural network for nuclide identification, TBNN (Li)~\cite{Li2024}. Finally, we introduce TBNN (ours), a more general transformer that modifies the Vision Transformer (ViT)~\cite{dosovitskiy2021image} for source identification in gamma spectroscopy (Fig.~\ref{fig:block_diagram}). Below, we motivate each architecture class by the assumptions it encodes about spectral data and the implications for generalization:

\begin{description}
    \item[Multilayer Perceptron] MLPs use one or more fully connected layers that project the input via learned weight matrices, apply a nonlinear activation function, and pass the result to subsequent layers. MLPs can be universal function approximators, and therefore have extremely high capacity for learning classification rules from data. Unfortunately, there are no inductive biases within MLPs, and they must encounter data examples of all realistic features. In the gamma spectroscopy domain, this manifests itself as MLPs struggling to generalize across unseen calibrations and resolutions.
    \item[Convolutional Neural Networks] CNNs incorporate one-dimensional convolutional layer(s) that slide learnable filters across local segments of the spectrum. To overcome issues with MLPs in a finite lattice domain (i.e., in spectra, where channels are equally spaced along one dimension, or images where pixels are equally spaced along two dimensions), CNNs assume that information from channels close to each other is correlated, and that the same feature will appear the same way in different spatial areas of the lattice. In gamma spectroscopy, this should enable generalization to shifts in calibration, although the \emph{a priori} choice of convolution size may be difficult to tune for detectors of different resolutions.
    \item[Transformers] For data representable as sequences, many different sequence models have been proposed and tested, progressing from RNNs~\cite{Goodfellow2016}, Transformers~\cite{Vaswani2017}, and recent selective state space models (SSMs)~\cite{guMambaLinearTimeSequence2023}. Transformers are the most common choice due to their ability to train in a fully parallel manner, and the amount of literature available characterizing them. In gamma spectroscopy, their attention mechanism should be able to determine higher order interactions between the number of counts in different channels, which is analogous to what gamma spectroscopists perform implicitly. Limitations in interaction range are often cited for Transformers, so widely separated peaks from a specific isotope may be difficult to capture.
    
    In TBNN (Li)~\cite{Li2024}, the authors first reshape a 1024-dimensional spectral input into a $32 \times 32$ sequence and then add fixed sinusoidal positional encodings. The resulting $32 \times 32$ embeddings are then fed into a stack of multi-head self-attention blocks, with a final dense layer mapping to the predicted isotopes. We introduce a more general transformer-based neural network, TBNN (ours), featuring several modifications over Li \emph{et al.}~\cite{Li2024}:
    \begin{enumerate}
        \item \textbf{Learnable patch embeddings} via a small CNN to map counts in each patch into embedding vectors, rather than using the counts directly. A simple linear layer or MLP could also be used.
        \item \textbf{Tunable patch size and embedding dimension} to optimize the trade-off between local detail preservation and computational efficiency.
        \item \textbf{Learnable [CLS] token} to aggregate information from all patches to produce the final prediction rather than simple flattening.
        \item \textbf{Learnable positional encoding} to allow the model to adapt the positional information based on the specific characteristics of the spectral data.
        \item \textbf{Pre-norm residual ordering} to improve training stability and convergence.
\end{enumerate}
\end{description}

\begin{figure}[t]
    \centering
    \adjustbox{max width=\linewidth}{
        \begin{tikzpicture}[
            node distance=1cm and 1.0cm,
            block/.style={rectangle, draw, thick, fill=blue!20, text width=6em, align=center, minimum height=4em},
            io/.style={ellipse, draw, thick, fill=green!20, align=center, minimum height=3em, text width=6em},
            plus_circle/.style={circle, draw, thick, fill=orange!30, minimum size=0.8cm},
            annot/.style={font=\small\itshape, text width=10em, align=center},
            connector/.style={-latex, thick},
            change/.style={draw, thick, rounded corners, fill=yellow!15, align=left, font=\footnotesize, text width=13em}
        ]
    
        \node[io] (input) {Input Spectrum};
    
        \node[block, right=of input] (patching) {Patch Embedding};
    
        \node[block, right=of patching] (cls_token) {Prepend [CLS] Token};
        \node[plus_circle, right=of cls_token] (add_pos) {+}; 
        \node[block, right=of add_pos] (pos_enc) {Positional Encoding};
    
        \node[block, below=2cm of add_pos] (ln1) {Layer Norm};
        \node[block, below=of ln1] (mha) {Multi-Head Attention};
        \node[plus_circle, below=of mha] (add1) {+};
    
        \node[block, right=of add1] (ln2) {Layer Norm};
        \node[block, below=of ln2] (ffn) {Feed-Forward Network};
        \node[plus_circle, below=of ffn] (add2) {+};
    
        \draw[thick, dashed]
            ([xshift=-0.3cm, yshift=0.7cm]ln1.north west)
            rectangle
            ([xshift=1.2cm, yshift=-0.7cm]add2.south east);
        \node[font=\bfseries, anchor=west, align=left]
              at ([xshift=3cm, yshift=-3cm]ln1.east)
              {Transformer\\Encoder\\Blocks};
    
        \node[block, below=1.5cm of add2] (extract_cls) {Extract [CLS] Token};
        \node[block, right=of extract_cls] (mlp_head) {MLP Head};
        \node[io, right=of mlp_head] (output) {Isotopic Prediction};
    
        \draw[connector] (input) -- (patching);
        \draw[connector] (patching) -- (cls_token);
        \draw[connector] (cls_token) -- (add_pos);
        \draw[connector] (pos_enc) -- (add_pos);
        \draw[connector] (add_pos) -- (ln1);
        \draw[connector] (ln1) -- (mha);
        \draw[connector] (mha) -- (add1);
        \draw[connector]
          (add_pos.south) -- ++(0,-0.7cm)
          -- ++(-2cm,0)
          -- ++(0,-2cm)
          |- (add1.west)
          node[pos=.25, above left, font=\itshape\small]{Residual};
        \draw[connector] (add1) -- (ln2);
        \draw[connector] (ln2) -- (ffn);
        \draw[connector] (ffn) -- (add2);
        \draw[connector] (add1.east) -- ++(0.5,0) |- (add2.west) node[pos=0.25, above left, font=\itshape\small] {Residual};
        \draw[connector] (add2) -- (extract_cls);
        \draw[connector] (extract_cls) -- (mlp_head);
        \draw[connector] (mlp_head) -- (output);
        
        \node[change, text width=19em, above=1cm of patching, anchor=south] (chg1)
            {(1) Learnable patch embeddings \\
             (2) Tunable patch size and embedding dimension};
        \draw[connector, dashed] (chg1.south) -- (patching.north);

        \node[change, text width=10em,
              below=6cm of cls_token,
              xshift=-2cm,
              anchor=south] (chg2)
            {(3) Learnable [CLS] token};
        
        \path let \p1 = (chg2.north) in
              coordinate (aux_up)   at (\x1, {\y1+3cm});
        \path let \p2 = (cls_token.south), \p3 = (aux_up) in
              coordinate (aux_left) at (\x2, \y3);
        
        \draw[connector, dashed]
            (chg2.north) -- (aux_up) -- (aux_left) -- (cls_token.south);
        \draw[connector, dashed] (chg2.south) |- (extract_cls.west);

        \node[change, above=1.5cm of pos_enc, anchor=south] (chg5)
            {(4) Learnable positional encoding};
        \draw[connector, dashed] (chg5.south) -- (pos_enc.north);

        \node[change, text width=12em, right=1.4cm of ln1, anchor=west] (chg4)
            {(5) Pre-norm residual ordering};
        \draw[connector, dashed] (chg4.west) -- (ln1.east);
        \draw[connector, dashed] (chg4.west) -- (ln2.north);
        
        \end{tikzpicture}
    }
    \caption{Proposed TBNN (ours) architecture with annotated modifications relative to Li \emph{et al.}~\cite{Li2024}}.
    \label{fig:block_diagram}
\end{figure}

\subsection{Training Protocols}

For each domain adaptation scenario, we trained three classes of models using the datasets generated in Section~\ref{Dataset Curation}. The first class, \emph{source-only}, was a single model simply trained on the entire source domain (\gadras{}) dataset. The second class of models, \emph{target-only}, were trained solely on the target-domain dataset (\geant{} in the sim-to-sim scenario; experiment in the sim-to-real scenario). Within this class, we trained multiple different models, each using a random subset of the target domain data of size \(2^i\) for \(i = 1,\dots,\lfloor\log_2 N\rfloor\), where \(N\) is the total number of training examples. The third class of models, \emph{domain-adapted}, were pretrained using the source-domain dataset, and subsequently fine-tuned using the target-domain data. As before, multiple domain-adapted models were trained with data subsets of varying sizes to evaluate the effect of dataset size on model performance. For all models, input spectra were first z-score normalized, which is a common data preprocessing technique in machine learning to stabilize training~\cite{Goodfellow-et-al-2016}.

\subsection{Hyperparameter Search}

Prior to training, we performed an in-depth Bayesian hyperparameter search using the Optuna software package~\cite{optuna2019}. We determined architecture hyperparameters using the source domain dataset in the sim-to-sim scenario with validation loss as our search criterion. To ensure a direct comparison, these architecture hyperparameters were used for all models in this study, while training hyperparameters (learning rate, batch size, dropout, weight decay) were calculated separately for each domain adaptation scenario. Since we might expect the optimal training hyperparameters to vary with dataset size, we performed a separate hyperparameter search for every data subset size. The fine-tuning hyperparameters were determined separately from the target-only hyperparameters, and the fine-tuning runs considered which (if any) layers to freeze as part of the hyperparameter search. Similar to the results of Ref.~\cite{shirokikh2020first}, we sometimes found that freezing the output layer(s) while fine-tuning only the earlier layers yielded better performance. We provide a further discussion of the hyperparameter tuning process in Section~\ref{hyperparameter_search}.

\subsection{Uncertainty Estimation}
\label{Uncertainty Estimation}

We employed an ensemble method to estimate the uncertainty in our model performance. Each model was trained with 10 different random initial configurations, and the variations in performance were used to quantify model uncertainty. For the source-only model class, this simply meant training 10 models for each architecture using different random weight initializations and seed configurations. For the target-only and domain-adapted model classes, we additionally sampled a random subset of target-domain data during each random trial. These trials were paired by ensuring that the random data subset was identical between target-only and domain-adapted models with the same seed. Similarly, each fine-tuning trial was initialized from the source-only pretraining checkpoint corresponding to its seed, so the measured variance reflects both pretraining and fine-tuning. Our reported uncertainty estimate reflects the empirical sample standard deviation across these 10 trials along with an added term that accounts for the sampling variability arising from a finite test set size.

\section{Results}
\label{Results}

\subsection{Predictive Performance (Classification Metrics)}
\label{predictive performance}

After training three classes of models (source-only, target-only, domain-adapted) for each of the three domain adaptation scenarios (sim-to-sim \HPGe, sim-to-real \LaBr, sim-to-real \NaI) with different architectural backbones (MLP, CNN, TBNN (Li), TBNN (ours)) for a range of different dataset sizes, we recorded the performance of each model on a held-out target-domain testing dataset. The choice of a best metric to be used for spectroscopic analysis has no consensus in the literature, although the detection limit and quantification uncertainty are preferred by spectroscopists for the detection and quantification of individual isotopes~\cite{Hagen2022, fernandezImprovementNuclideDetection2025, Hagen2021}. Without selecting details about the amount of background present and the isotopes of interest, these metrics are not computable, and traditional machine learning metrics characterizing the classifiers themselves must be used\footnote{Unfortunately, some evidence in~\cite{fernandezImprovementNuclideDetection2025} shows that accuracy may not be monotonic with detection limit or quantification accuracy, so caution should be used when extending classifier metrics to real-world performance.}. For the sim-to-real scenarios, we used testing accuracy as our evaluation metric. For the sim-to-sim scenario, accuracy is no longer a well-defined metric since we are performing multi-label proportion estimation; instead, we computed the testing absolute proportion error (APE) score as our evaluation metric. The APE score (Eq.~\ref{eq:APE_score}) provides a convenient 0-to-1 scoring metric to compare how precisely different models predict the fractional composition of each mixed source. We provide a more detailed discussion of the APE score as an evaluation metric in Section~\ref{evaluation_metric}.

\subsubsection{Sim-to-sim comparison}
\label{sim-to-sim comparison}

We summarize the results of the sim-to-sim scenario in Fig.~\ref{fig:combined_ape_hypothesis}, where we compare the \geant{} APE score across different model classes for each architecture. Overall, we observe that the APE scores of the domain-adapted models almost always match or exceed those of the source-only and target-only models. In particular, using a fine-tuning dataset of 1024 spectra, the TBNN (ours) model achieves a testing APE score of $0.762 \pm 0.002$, outperforming both the source-only model (APE score = $0.652 \pm 0.003$) and the target-only model (APE score = $0.554 \pm 0.012$) on the same dataset size. We quantify these results in Table~\ref{subtable:sim-to-sim-HPGe} where we explicitly record the testing APE scores for models trained with 0 versus 64 versus 1024 target domain training spectra.

To test the statistical significance of these conclusions, we performed a series of one-sided Wilcoxon signed-rank tests using the output of 10 paired trials with random initial configurations, as described in Section~\ref{Uncertainty Estimation}. In these tests, the null hypothesis states that the difference in APE scores between the domain-adapted models and the baseline models (source-only or target-only) is nonpositive, and we use a significance threshold of $\alpha=0.01$. These results are indicated in the bottom panels of Fig.~\ref{fig:combined_ape_hypothesis}, where we can reject the null hypothesis for nearly all cases, except domain-adapted versus source-only at small fine-tuning sizes.

\subsubsection{Sim-to-real comparison}
\label{sim-to-real comparison}

We perform the same analysis for the sim-to-real comparisons and summarize the results in Figs~\ref{fig:combined_accuracy_hypothesis_LaBr} and~\ref{fig:combined_accuracy_hypothesis_NaI}, comparing the testing accuracy across different model classes and architectures. These results are quantified in Tables~\ref{subtable:sim-to-real-LaBr} and~\ref{subtable:sim-to-real-NaI}. Similar to the sim-to-sim scenario, we observe a statistically significant improvement in testing accuracy for most training sizes. In particular, using 64 fine-tuning spectra, the TBNN (ours) model achieves an accuracy of $0.955 \pm 0.025$ on the {\LaBr} testing dataset, easily surpassing the source-only model (accuracy = $0.745 \pm 0.010$) and the target-only model (accuracy = $0.795 \pm 0.047$) trained on the same experimental subset. We see a similar result on the {\NaI} dataset, achieving an accuracy of $0.946 \pm 0.032$ using a TBNN (ours) architecture and 64 fine-tuning spectra, versus $0.738 \pm 0.024$ of the source-only model and $0.785 \pm 0.043$ of the target-only model.

\begin{table}
\sisetup{
    separate-uncertainty,
    retain-zero-uncertainty,
    table-format=1.3(3)
}
\centering
\caption{Performance comparison of source-only, target-only, and domain-adapted models, trained using a variable number of spectra (\textit{N}) from the target dataset. Cell values represent the mean and sample standard deviation of the indicated performance metric from 10 repeated trials. Background color intensity follows a non-linear scale to enhance contrast.
}
\label{table:performance_comparison}
\begin{subtable}{\textwidth}
    \centering
    \caption{Scenario: sim-to-sim (\HPGe). Cell entries indicate testing APE score}
    \label{subtable:sim-to-sim-HPGe}
    \begin{tabular}{ l S S S S S }
    \toprule
    & \multicolumn{2}{c}{Target-only} & {Source-only} & \multicolumn{2}{c}{Domain-adapted} \\
    \cmidrule(lr){2-3} \cmidrule(lr){4-4} \cmidrule(lr){5-6}
    & {\textit{N}=64} & {\textit{N}=1024} & {\textit{N}=0} & {\textit{N}=64} & {\textit{N}=1024} \\
    \midrule
    MLP  
    & \colorcell{0.252}{0.252 \pm 0.003}
    & \colorcell{0.543}{0.543 \pm 0.003}
    & \colorcell{0.612}{0.612 \pm 0.001}
    & \colorcell{0.639}{0.639 \pm 0.008}
    & \colorcell{0.713}{0.713 \pm 0.002}
    \\
    CNN  
    & \colorcell{0.303}{0.303 \pm 0.009}
    & \colorcell{0.556}{0.556 \pm 0.004}
    & \colorcell{0.635}{0.635 \pm 0.003}
    & \colorcell{0.585}{0.585 \pm 0.016}
    & \colorcell{0.696}{0.696 \pm 0.003}
    \\
    TBNN (Li)
    & \colorcell{0.284}{0.284 \pm 0.010}
    & \colorcell{0.600}{0.600 \pm 0.005}
    & \colorcell{0.635}{0.635 \pm 0.005}
    & \colorcell{0.691}{0.691 \pm 0.005}
    & \colorcell{0.767}{0.767 \pm 0.003}
    \\
    TBNN (ours)
    & \colorcell{0.228}{0.228 \pm 0.007}
    & \colorcell{0.554}{0.554 \pm 0.012}
    & \colorcell{0.652}{0.652 \pm 0.003}
    & \colorcell{0.700}{0.700 \pm 0.006}
    & \colorcell{0.762}{0.762 \pm 0.002}
    \\
    \bottomrule
    \end{tabular}
\end{subtable}

\vspace{1.5em}
\begin{subtable}{\textwidth}
    \centering
    \caption{Scenario: sim-to-real (\LaBr). Cell entries indicate testing accuracy}
    \label{subtable:sim-to-real-LaBr}
    \begin{tabular}{ l S S S S S } 
    \toprule
    & \multicolumn{2}{c}{Target-only} & {Source-only} & \multicolumn{2}{c}{Domain-adapted} \\
    \cmidrule(lr){2-3} \cmidrule(lr){4-4} \cmidrule(lr){5-6}
    & {\textit{N}=64} & {\textit{N}=1024} & {\textit{N}=0} & {\textit{N}=64} & {\textit{N}=1024} \\
    \midrule
    MLP  
    & \colorcell{0.765}{0.765 \pm 0.052}
    & \colorcell{1.000}{1.000 \pm 0.000}
    & \colorcell{0.741}{0.741 \pm 0.009}
    & \colorcell{0.965}{0.965 \pm 0.017}
    & \colorcell{1.000}{1.000 \pm 0.000}
    \\
    CNN  
    & \colorcell{0.781}{0.781 \pm 0.048}
    & \colorcell{1.000}{1.000 \pm 0.000}
    & \colorcell{0.742}{0.742 \pm 0.011}
    & \colorcell{0.957}{0.957 \pm 0.018}
    & \colorcell{1.000}{1.000 \pm 0.000}
    \\
    TBNN (Li)
    & \colorcell{0.757}{0.757 \pm 0.054}
    & \colorcell{1.000}{1.000 \pm 0.000}
    & \colorcell{0.763}{0.763 \pm 0.011}
    & \colorcell{0.958}{0.958 \pm 0.019}
    & \colorcell{1.000}{1.000 \pm 0.000}
    \\
    TBNN (ours)
    & \colorcell{0.795}{0.795 \pm 0.047}
    & \colorcell{1.000}{1.000 \pm 0.000}
    & \colorcell{0.745}{0.745 \pm 0.010}
    & \colorcell{0.955}{0.955 \pm 0.025}
    & \colorcell{1.000}{1.000 \pm 0.000}
    \\
    \bottomrule
    \end{tabular}
\end{subtable}

\vspace{1.5em}
\begin{subtable}{\textwidth}
    \centering
    \caption{Scenario: sim-to-real (\NaI). Cell entries indicate testing accuracy}
    \label{subtable:sim-to-real-NaI}
    \begin{tabular}{ l S S S S S } 
    \toprule
    & \multicolumn{2}{c}{Target-only} & {Source-only} & \multicolumn{2}{c}{Domain-adapted} \\
    \cmidrule(lr){2-3} \cmidrule(lr){4-4} \cmidrule(lr){5-6}
    & {\textit{N}=64} & {\textit{N}=1024} & {\textit{N}=0} & {\textit{N}=64} & {\textit{N}=1024} \\
    \midrule
    MLP  
    & \colorcell{0.767}{0.767 \pm 0.049}
    & \colorcell{1.000}{1.000 \pm 0.000}
    & \colorcell{0.738}{0.738 \pm 0.012}
    & \colorcell{0.940}{0.940 \pm 0.042}
    & \colorcell{1.000}{1.000 \pm 0.000}
    \\
    CNN  
    & \colorcell{0.766}{0.766 \pm 0.056}
    & \colorcell{1.000}{1.000 \pm 0.000}
    & \colorcell{0.755}{0.755 \pm 0.012}
    & \colorcell{0.955}{0.955 \pm 0.035}
    & \colorcell{1.000}{1.000 \pm 0.000}
    \\
    TBNN (Li)
    & \colorcell{0.762}{0.762 \pm 0.052}
    & \colorcell{1.000}{1.000 \pm 0.000}
    & \colorcell{0.753}{0.753 \pm 0.013}
    & \colorcell{0.955}{0.955 \pm 0.021}
    & \colorcell{1.000}{1.000 \pm 0.000}
    \\
    TBNN (ours)
    & \colorcell{0.785}{0.785 \pm 0.043}
    & \colorcell{1.000}{1.000 \pm 0.000}
    & \colorcell{0.738}{0.738 \pm 0.024}
    & \colorcell{0.946}{0.946 \pm 0.032}
    & \colorcell{1.000}{1.000 \pm 0.000}
    \\
    \bottomrule
    \end{tabular}
\end{subtable}

\end{table}

\subsubsection{Transformer vs. Classical Architectures}

To quantify architectural differences in model performance, we computed each model's testing evaluation metric for all fine-tuning dataset sizes. For each dataset size, we performed two-sided Wilcoxon signed-rank tests to test the null hypothesis that there is no difference in testing score between each model architecture. In the sim-to-sim scenario (Fig.~\ref{fig:rank_geant}), the transformer architectures (TBNN (Li) and TBNN (ours)) perform strongly across most sizes: TBNN (ours) is statistically superior at small sizes ($2^1$-$2^7$), while TBNN (Li) is best at intermediate sizes ($2^7$-$2^{17}$). The MLP only leads at large sizes (${\geq} 2^{16}$), and the CNN does not lead at any size. We quantify these results in Table~\ref{table:pvalue_arch_comparison_hpge}, where overall, domain-adapted transformer architectures demonstrate a clear statistical advantage over classical architectures across a wide range of fine-tuning sizes. Notably, these trends do not appear in the target-only model class, suggesting that transformers benefit more from pretraining than classical architectures. We hypothesize that this result reflects the higher model complexity and greater data requirements of transformers, and that rigorous synthetic pretraining can partially mitigate this large-data hurdle.

In the sim-to-real scenarios (Figs.~\ref{fig:rank_nai} and~\ref{fig:rank_labr}), we still observe a statistically significant performance gain from transformers at some dataset sizes. However, the contrast is far less widespread, as most dataset sizes show no statistically significant difference between model architectures. This result suggests that, for this particular sim-to-real task, the classification boundary is not sufficiently complex for more expressive architectures (TBNNs) to provide a consistent advantage over classical architectures (MLPs, CNNs). We quantify these results in Tables~\ref{table:pvalue_arch_comparison_labr} and~\ref{table:pvalue_arch_comparison_nai}, where we find that transformers retain a slight, but statistically significant, edge over classical architectures for some, but not all, dataset sizes. Overall, we recommend that future work repeat such architectural comparisons on more challenging experimental classification scenarios.

\subsection{Model Analysis (Diagnostics \& Explainability)}
\label{model analysis}

In Section~\ref{predictive performance}, we present statistical evidence that fine-tuning yields an improvement to model performance as measured by classification metrics (accuracy, APE score). However, a comparison between the accuracy of two models offers little insight as to \emph{why} one model outperforms the other, or whether such improvement would necessarily transfer to a new research scenario. To enrich this analysis, we include comparisons across several diagnostic metrics (calibration, separation, and smoothness), as well as qualitative understanding using SHapley Additive exPlanations (SHAP).

\subsubsection{Model Diagnostics}
\label{model diagnostics}
In addition to classification accuracy, we consider a range of well-established metrics from literature that quantify calibration, separation, uncertainty, and input-space smoothness. Calibration quantifies a model's confidence in its predictions, penalizing the gap between a model's predicted confidence and its actual accuracy. Separation evaluates how decisively a model makes its predictions by comparing the confidence in the chosen class to the next most likely class. Uncertainty measures a model's confusion between different classes, penalizing scenarios where a model's prediction is split among several classes. Smoothness assesses the model's decision boundary, rewarding models that make similar predictions for similar inputs. Our goal with these comparisons is to include grounded, complementary techniques to offer insights into model performance beyond pure classification accuracy.

We present our results in Table~\ref{table:metric-comparison} using a TBNN (ours) architecture for the sim-to-real {\LaBr} adaptation scenario. In addition to standard classification accuracy ($\texttt{acc}$), we include negative log-likelihood ($\texttt{nll}$), which more heavily penalizes overconfidence~\cite{murphy2012machine}. The Brier score ($\texttt{brier}$) measures the mean-squared error between the true one-hot encoded label and predicted probability vectors~\cite{glenn1950verification}, while the expected calibration error ($\texttt{ece}$) bins predictions by confidence to directly measure the confidence-accuracy gap~\cite{guo2017calibration}. Prediction margin evaluates the mean ($\texttt{margin\_mean}$) and 10th percentile ($\texttt{margin\_p10}$) of the margin for each spectrum, defined as the difference in predicted probabilities between the top two classes~\cite{bartlett1998boosting}. Mean predictive entropy ($\texttt{entropy\_mean}$) measures how confident or uncertain the model is in its predictions by calculating the average Shannon entropy of the predicted probability vectors~\cite{shannon1948mathematical}. Mean Jacobian norm ($\texttt{jacobian\_norm\_mean}$) quantifies a model's sensitivity to its input by averaging the squared $\ell_2$-norm of the gradient of the loss with respect to the input across all samples~\cite{goodfellow2014explaining}. Lastly, we considered four boundary complexity and smoothness metrics by first building a $k$-nearest neighbors graph using a Euclidean distance metric and $k=10$ on the input $\ell_1$-normalized spectra of the testing dataset. We considered a range of distance metrics (Euclidean, cosine, correlation) and different values for $k$ (5-30), finding minimal changes in results. We then considered the post-softmax probability vectors $(p_u, p_v)$ for each edge and computed the following metrics~\cite{belkin2006manifold,zhu2003semi}:

\begin{itemize}
  \item \texttt{knn\_tv\_hard}: fraction of pairs with different predicted hard labels (i.e., how often \(\arg\max p_u \neq \arg\max p_v \)).
  \item \texttt{knn\_prob\_l2}: mean squared \(\ell_2\) distance between probability vectors (i.e., \(\lVert p_u - p_v \rVert_2^2\)).
  \item \texttt{knn\_conf\_absdiff}: average absolute change in confidence between probability vectors (i.e., \(\lvert \max p_u - \max p_v \rvert\)).
  \item \texttt{knn\_margin\_absdiff}: average absolute change in true-class margin (i.e., define \(m_i = \log p_{i,y_i} - \max_{c\ne y_i} \log p_{i,c}\), where $y_i$ is the ground-truth class for sample $i$. Then, take \(\lvert m_u - m_v \rvert\)).
\end{itemize}

To test the statistical significance of each diagnostic metric, we ran a one-sided Wilcoxon signed-rank test comparing the empirically best model to the second best model across 10 paired random trials. The results of this section are summarized in Table~\ref{table:metric-comparison}, where we see that the source-only model wins in 1 out of 12 metrics, target-only wins in 2 out of 12 metrics, and domain-adapted in 7 out of 12 metrics, with 2 out of 12 metrics considered statistically insignificant. In particular, the domain-adapted model performs well in the calibration, separation, and sensitivity metrics, suggesting more trustworthy predictions. The input-space smoothness metrics yield mixed results, suggesting that fine-tuning primarily improves reliability as opposed to dramatically altering the input-space geometry of the decision boundary.

\begin{table}
\centering
\caption{TBNN (ours) model comparisons using different diagnostic metrics ($\uparrow$ higher is better, $\downarrow$ lower is better) on the {\LaBr} dataset with 64 experimental training spectra. Statistical significance is tested using a one-sided Wilcoxon signed-rank test, and entries for the best model are bolded when significant ($\alpha < 0.01$). Overall, fine-tuning provides improvements using a wide variety of metrics beyond classification accuracy.}
\adjustbox{max width=\linewidth}{
    \begin{tabular}{
      l
      l 
      l 
      l 
      c 
      c 
      l 
    }
    \toprule
    \textbf{Metric} & \textbf{Source-only} & \textbf{Target-only 64} & \textbf{Domain-adapted 64} & \makecell{$p$\textbf{-value}\\[-0.5ex]\textbf{(best vs. 2nd)} } & \textbf{Dir.} & \textbf{Type} \\
    \midrule
    \texttt{acc} & $0.745 \pm 0.010$ & $0.795 \pm 0.047$ & $\mathbf{0.955 \pm 0.025}$ & $\mathbf{0.001}$ & $\uparrow$ & Accuracy \\
    \texttt{nll}~\cite{murphy2012machine} & $2.91 \pm 0.34$ & $1.18 \pm 0.25$ & $\mathbf{0.284 \pm 0.201}$ & $\mathbf{0.001}$ & $\downarrow$ & Likelihood / Loss \\
    \texttt{brier}~\cite{glenn1950verification} & $0.488 \pm 0.010$ & $0.292 \pm 0.061$ & $\mathbf{0.075 \pm 0.042}$ & $\mathbf{0.001}$ & $\downarrow$ & Calibration \\
    \texttt{ece}~\cite{guo2017calibration} & $0.239 \pm 0.006$ & $0.071 \pm 0.011$ & $\mathbf{0.027 \pm 0.016}$ & $\mathbf{0.001}$ & $\downarrow$ & Calibration \\
    \texttt{margin\_mean}~\cite{bartlett1998boosting} & $6.57 \pm 0.29$ & $2.19 \pm 0.49$ & $\mathbf{7.82 \pm 0.91}$ & $\mathbf{0.002}$ & $\uparrow$ & Separation / Margin \\
    \texttt{margin\_p10}~\cite{bartlett1998boosting} & $-13.3 \pm 1.3$ & $-4.70 \pm 1.14$ & $\mathbf{3.14 \pm 1.50}$ & $\mathbf{0.001}$ & $\uparrow$ & Separation / Margin \\
    \texttt{entropy\_mean}~\cite{shannon1948mathematical} & $\mathbf{0.057 \pm 0.015}$ & $0.753 \pm 0.101$ & $0.117 \pm 0.088$ & $\mathbf{0.007}$ & $\downarrow$ & Uncertainty \\
    \texttt{jacobian\_norm\_mean}~\cite{goodfellow2014explaining} & $6.56 \pm 1.35$ & $4.20 \pm 1.92$ & $\mathbf{1.26 \pm 0.92}$ & $\mathbf{0.001}$ & $\downarrow$ & Sensitivity \\
    \texttt{knn\_tv\_hard}~\cite{belkin2006manifold,zhu2003semi} & $0.960 \pm 0.001$ & $\mathbf{0.949 \pm 0.006}$ & $0.959 \pm 0.003$ & $\mathbf{0.001}$ & $\downarrow$ & Boundary Complexity \\
    \texttt{knn\_prob\_l2}~\cite{belkin2006manifold,zhu2003semi} & $1.90 \pm 0.01$ & $\mathbf{1.38 \pm 0.09}$ & $1.87 \pm 0.03$ & $\mathbf{0.001}$ & $\downarrow$ & Smoothness \\
    \texttt{knn\_conf\_absdiff}~\cite{belkin2006manifold,zhu2003semi} & $0.017 \pm 0.005$ & $0.229 \pm 0.030$ & $0.033 \pm 0.018$ & $0.010$ & $\downarrow$ & Smoothness \\
    \texttt{knn\_margin\_absdiff}~\cite{belkin2006manifold,zhu2003semi} & $7.51 \pm 0.47$ & $3.91 \pm 0.49$ & $3.75 \pm 0.73$ & $0.278$ & $\downarrow$ & Smoothness \\
    \bottomrule
    \end{tabular}
}
\label{table:metric-comparison}
\end{table}

\subsubsection{Model Interpretability}
\label{model interpretability}

We also present evidence that beyond metric scores, domain-adapted models learn more human-interpretable features compared to target-only models. In Explainable AI literature, various techniques have been proposed for quantifying input feature relevance, such as gradient-based~\cite{simonyan2014deepinsideconvolutionalnetworks, smilkov2017smoothgradremovingnoiseadding} and Attention Rollout~\cite{abnar2020quantifyingattentionflowtransformers} methods. In Ref.~\cite{bandstra2023explaining}, the authors considered various explanation methods in the context of gamma spectroscopy, and concluded that SHAP offered especially accurate explanations of radioisotope predictions compared to other methods. In short, SHAP evaluates an input feature's importance by removing the feature and assessing the resulting effect on the model's output. A more mathematically thorough description of the KernelSHAP algorithm can be found in Refs.~\cite{bandstra2023explaining, lundberg2017unifiedapproachinterpretingmodel}.

\begin{figure}
    \centering
    \includegraphics[width=0.48\textwidth]{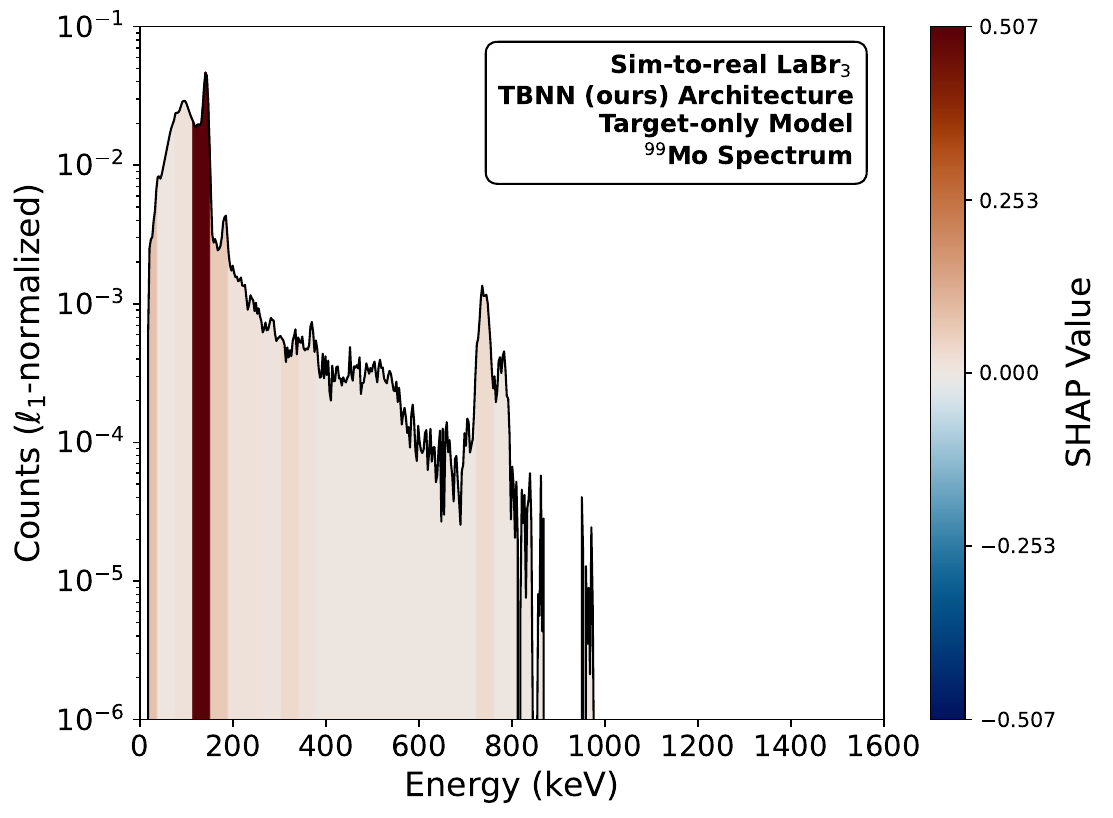}
    \includegraphics[width=0.48\textwidth]{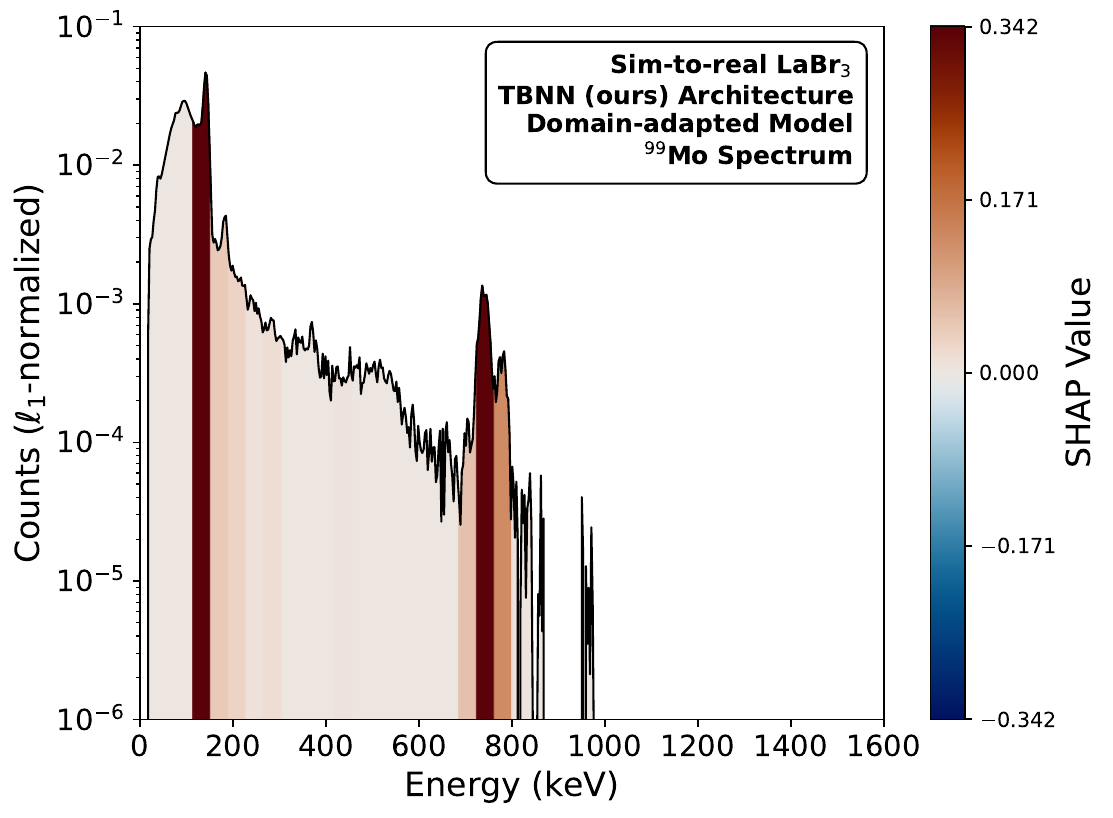}
    \caption{SHAP explanations for a $^{99}$Mo spectrum using a target-only model (left) and a domain-adapted model (right). The target-only model only identifies the 140.5 keV line from \textsuperscript{99m}Tc as salient, whereas the domain-adapted model also identifies the 739.5 keV peak characteristic of $^{99}$Mo.}
    \label{fig:XAI_Mo99}
\end{figure}

For this analysis, we chose a target-only and domain-adapted TBNN (ours) model from the sim-to-real {\LaBr} adaptation scenario, each trained with the same 64 experimental spectra. Fig.~\ref{fig:XAI_Mo99} shows a $^{99}$Mo spectrum overlaid with a color plot of calculated SHAP explanation values for each region. Large, positive SHAP values (highlighted in red) indicate that a feature positively correlates with the model's class prediction. The SHAP explanation reveals that the target-only model relies exclusively on the 140.5 keV peak from \textsuperscript{99m}Tc when computing its classification, ignoring the 739.5 keV peak characteristic of $^{99}$Mo. As a result, the target-only model frequently confuses \textsuperscript{99m}Tc and $^{99}$Mo due to reliance on the shared daughter line, resulting in an overall false alarm rate of 2.0\% for $^{99}$Mo. Conversely, the domain-adapted model correctly identifies both the 140.5 keV and 739.5 keV peaks, enabling unambiguous discrimination between both isotopes. We report similar findings using other architectures and include the same $^{99}$Mo SHAP comparisons for MLP, CNN, and TBNN (Li) architectures in Fig.~\ref{fig:XAI_Mo99_all}.

Our analysis further reveals that target-only models will often overfit to the low energy region of the spectrum. As seen in Fig.~\ref{fig:XAI_Pu239}, when classifying $^{239}$Pu, the target-only model identifies the $20$-$40$ keV band as the most salient component of the spectrum, indicative of overfitting. Conversely, the domain-adapted model's SHAP explanation most strongly highlights a region containing the 345, 375, and 413 keV peaks, which are commonly used by spectroscopists to identify $^{239}$Pu, especially in the presence of shielding~\cite{reilly1991passive}. Overall, we observe that domain-adapted models more consistently produce explanations that align with human-interpretable photopeaks, while target-only models are more likely to focus on low-energy, high-count features. As such, we hypothesize that \gadras{} pretraining confers an interpretability boost, even when the target-only and domain-adapted models have similar per-class accuracy.

We conclude our analysis by examining the behavior of neural network classification models when different isotopes exhibit shared spectral features. Using the domain-adapted TBNN (ours) model, we computed SHAP explanations for a $^{18}$F and $^{22}$Na spectrum, shown in Fig.~\ref{fig:XAI_F18_Na22}. The model identifies the 511 keV positron–electron annihilation peak as salient in both spectra; however, the model also highlights the 1274.5 keV line in the $^{22}$Na spectrum. This result provides further evidence that the model has learned to utilize secondary features to distinguish between isotopes with shared energy peaks.

\section{Conclusion}
\label{Conclusion}

In this study, we present a supervised domain adaptation framework for fine-tuning models pretrained on synthetic data to perform classification in an experimental downstream setting. We consider both sim-to-sim and sim-to-real adaptation scenarios, and find that domain-adapted models provide a statistically significant improvement in classification metrics compared to synthetic-only models or experiment-only models with no pretraining. For example, using a custom transformer-based neural network implementation, we achieve a testing accuracy of 96\% after fine-tuning on 64 {\LaBr} spectra, improved from 75\% of a source-only model and 80\% using a target-only model. Furthermore, we find that domain-adapted models identify more human-interpretable features compared to target-only models. These results highlight a path forward to apply machine learning techniques to experimental settings where only a limited amount of experimental training data is available.

\section{Acknowledgements}

This research was supported by the Laboratory Directed Research and Development Program at Pacific Northwest National Laboratory, a multiprogram national laboratory operated by Battelle for the U.S. Department of Energy under contract DE-AC05-76RLO1830. Peter Lalor is grateful for the support of the Linus Pauling Distinguished Postdoctoral Fellowship. The authors would like to acknowledge Tyler Morrow and Brian Archambault for their useful suggestions and feedback. The authors declare no conflict of interest.



\newpage

\setlength{\bibsep}{0pt}
\bibliography{References.bib}

\newpage

\appendix

\section{Supplementary Materials}
\label{appendix}

\subsection{Defining an evaluation metric for mixed proportion estimation}
\label{evaluation_metric}

Choosing an evaluation metric for multi-label proportion estimation is less intuitive than for traditional classification problems. In multi-label proportion estimation scenarios, several isotopes may be present simultaneously, with the fractions of all isotopes summing to unity, rendering a simple metric such as accuracy elusive. Instead, we define the absolute proportion error (APE) score, computed by rescaling the mean absolute error to yield a value between 0 and 1. Explicitly,

\begin{gather}
\text{APE score} \coloneqq 1 - \frac{1}{2N} \sum_{i=1}^N \sum_{j=1}^M \lvert y_{\text{pred}, i, j} - y_{\text{true}, i, j} \rvert \label{eq:APE_score} \\
\sum_{j=1}^M y_{\text{pred}, i, j} = 1~,~~~\sum_{j=1}^M y_{\text{true}, i, j} = 1 ~.\nonumber
\end{gather}

In Eq.~\ref{eq:APE_score}, $N$ is the number of spectra in the testing dataset, $M$ is the number of radioisotope classes, and $y_\text{pred}$ and $y_\text{true}$ are $N \times M$ matrices of the predicted and true proportion labels, respectively. An APE score of 1 indicates perfect agreement between the predicted and true isotope proportions, while a score of 0 reflects maximal deviation. This evaluation metric provides a convenient way to compare how precisely different models predict the fractional composition of each mixed source.

\subsection{Hyperparameter search}
\label{hyperparameter_search}

For each architecture (MLP, CNN, TBNN (Li), TBNN (ours)), we conducted a Bayesian hyperparameter search with 150 trials on the sim-to-sim source domain dataset. Each model was trained for 20 minutes on an NVIDIA Tesla V100 GPU with 32 GB of memory, with the number of epochs adjusted dynamically to maintain a consistent overall training time. Underperforming configurations were pruned early, and an early stopping criterion was applied to mitigate overfitting. Hyperparameters were ranked by their validation cross-entropy loss, and the best-performing parameters are shown in Table~\ref{table:source_hps}. Source-only baseline models for all scenarios and all architectures were trained for 80 minutes on the same hardware, and the training was repeated 10 times using different random seeds. All trained models were compact ($<$~2~GB) and achieved single-threaded CPU inference latencies of $\lesssim$~2ms, indicating that they could easily be deployed on a commodity system.

When training the target-only and domain-adapted models, the same architecture hyperparameters were used as before, but training hyperparameters (learning rate, batch size, dropout, weight decay) were recalculated using a subsequent Bayesian hyperparameter search with 50 trials and 5 minutes per trial. The fine-tuning runs additionally included which layers to freeze (if any) as part of the hyperparameter search. We conducted a separate hyperparameter search for each scenario as well as for every data subset size. Table~\ref{table:all_scenarios_hps} lists the best hyperparameters for data subsets of size 64 and 1024 for the target-only and domain-adapted runs for each of the three domain adaptation scenarios.

\begingroup
\linespread{1.1}\selectfont
\begin{table}
\centering
\caption{Source-only hyperparameter optimization summary for MLP, CNN, TBNN (Li), and TBNN (ours) architectures. The table details the search spaces and best-performing hyperparameter values obtained via a Bayesian search using the source dataset from the sim-to-sim scenario. The best-performing architecture-specific hyperparameters are used for all models across different scenarios in this study. Conversely, the training-specific hyperparameters are only used for training the source-only models, and are subsequently recalculated for other training runs in Table~\ref{table:all_scenarios_hps}.}
\begin{tabular}{l l l l}
\toprule
Architecture & Parameter           & Search Space      & Best run \\ 
\midrule
MLP     & Learning Rate            & \{1e-5, 2e-3\}    & 5.12e-5 \\
        & Batch Size               & \{32, 512\}       & 512 \\
        & Weight Decay             & \{1e-7, 1e-1\}    & 2.67e-5 \\
        & Dropout                  & \{0.0, 0.4\}      & 0.346 \\
        & Num Dense Layers         & \{1, 4\}          & 2 \\
        & Dense1 Hidden Units      & \{512, 8192\}     & 4096 \\
        & Dense2 Hidden Units      & \{256, 4096\}     & 2048 \\
\midrule
CNN     & Learning Rate            & \{1e-5, 2e-3\}    & 1.17e-4 \\
        & Batch Size               & \{32, 512\}       & 512 \\
        & Weight Decay             & \{1e-7, 1e-1\}    & 3.42e-5 \\
        & Dropout                  & \{0.0, 0.4\}      & 0.101 \\
        & Num Convolutional Layers & \{1, 3\}          & 1 \\
        & Conv Filters             & \{16, 256\}       & 32 \\
        & Conv Kernel Size         & \{3, 9\}          & 7 \\
        & Num Dense Layers         & \{1, 2\}          & 2 \\
        & Dense1 Hidden Units      & \{512, 8192\}     & 2048 \\
        & Dense2 Hidden Units      & \{256, 4096\}     & 1024 \\
\midrule
TBNN (Li) & Learning Rate          & \{1e-5, 2e-3\}    & 1.02e-3 \\
        & Batch Size               & \{32, 512\}       & 256 \\
        & Weight Decay             & \{1e-7, 1e-1\}    & 3.46e-3 \\
        & Dropout                  & \{0.0, 0.4\}      & 4.53e-4 \\
        & Num Attention Blocks     & \{1, 8\}          & 5 \\
        & Num Heads                & \{1, 8\}          & 4 \\
        & FF Dimension             & \{64, 8192\}      & 1024 \\
\midrule
TBNN (ours) & Learning Rate        & \{1e-5, 2e-3\}    & 1.49e-4 \\
        & Batch Size               & \{32, 512\}       & 64 \\
        & Weight Decay             & \{1e-7, 1e-1\}    & 5.02e-3 \\
        & Dropout                  & \{0.0, 0.4\}      & 0.0198 \\
        & Embedding Method         & \{Linear, MLP, CNN\} & CNN \\
        & CNN Embed Filters        & \{8, 256\}        & 8 \\
        & Embedding Dimension      & \{8, 1024\}       & 256 \\
        & Num Attention Blocks     & \{1, 5\}          & 4 \\
        & Num Heads                & \{1, 8\}          & 8 \\
        & FF Dimension             & \{2, 4096\}       & 512 \\
        & Patch Size               & \{16, 64\}        & 64 \\
        & Positional Encoding      & \{Sinusoidal, Learnable\} & Learnable \\
\bottomrule
\end{tabular}
\label{table:source_hps}
\end{table}
\endgroup

\begingroup
\linespread{1.1}\selectfont
\begin{table}
\centering
\caption{Target-only and domain-adapted hyperparameter optimization summary for MLP, CNN, TBNN (Li), and TBNN (ours) architectures. The table details the search spaces and best-performing training hyperparameter values obtained via a Bayesian search for all domain adaptation scenarios: (a) sim-to-sim (\HPGe), (b) sim-to-real (\LaBr), and (c) sim-to-real (\NaI). For brevity, we only include results using data subsets of size $N=64$ and $N=1024$. The `Frozen Layers' parameter refers to individual trainable layers, not composite blocks.}
\label{table:all_scenarios_hps}

\adjustbox{max width=0.6\textwidth, max height=0.7\textheight}{
\begin{minipage}{\textwidth}

\begin{subtable}{\textwidth}
\centering
\caption{Training hyperparameters for the sim-to-sim \gadras{}$\to$\geant{} adaptation using a {\HPGe} detector.}
\label{table:sim2sim_hps}
\begin{tabular}{l l l l l l l}
\toprule
& & & \multicolumn{2}{c}{Best run (target-only)} & \multicolumn{2}{c}{Best run (domain-adapted)} \\
\cmidrule(lr){4-5} \cmidrule(lr){6-7}
Architecture & Parameter & Search Space & {\textit{N}=64} & {\textit{N}=1024} & {\textit{N}=64} & {\textit{N}=1024} \\
\midrule
MLP         & Learning Rate    & \{1e-6, 1e-3\}    & 9.58e-5     & 9.01e-4     & 2.00e-4        & 3.82e-4 \\
            & Batch Size       & \{32, 512\}       & 32          & 512         & 64             & 512 \\
            & Weight Decay     & \{1e-7, 1e-1\}    & 4.08e-4     & 2.92e-3     & 3.17e-2        & 3.69e-5 \\
            & Dropout          & \{0.0, 0.4\}      & 0.373       & 0.337       & 0.397          & 0.395 \\
            & Frozen Layers    & \{none, all\}     & N/A         & N/A         & None           & None \\
\midrule
CNN         & Learning Rate    & \{1e-6, 1e-3\}    & 1.92e-5     & 7.16e-4     & 9.68e-4        & 8.71e-5 \\
            & Batch Size       & \{32, 512\}       & 32          & 512         & 64             & 64 \\
            & Weight Decay     & \{1e-7, 1e-1\}    & 7.76e-2     & 2.84e-4     & 9.84e-7        & 5.36e-7 \\
            & Dropout          & \{0.0, 0.4\}      & 0.381       & 0.275       & 0.303          & 0.322 \\
            & Frozen Layers    & \{none, all\}     & N/A         & N/A         & Last 3 layers  & First 2 layers \\
\midrule
TBNN (Li)   & Learning Rate    & \{1e-6, 1e-3\}    & 6.73e-4     & 9.75e-4     & 6.44e-4        & 6.53e-4 \\
            & Batch Size       & \{32, 512\}       & 64          & 128         & 32             & 512 \\
            & Weight Decay     & \{1e-7, 1e-1\}    & 2.73e-7     & 2.14e-2     & 1.71e-5        & 6.86e-5 \\
            & Dropout          & \{0.0, 0.4\}      & 0.149       & 0.067       & 0.027          & 0.021 \\
            & Frozen Layers    & \{none, all\}     & N/A         & N/A         & None           & None \\
\midrule
TBNN (ours) & Learning Rate    & \{1e-6, 1e-3\}    & 3.37e-4     & 7.67e-4     & 1.82e-4        & 5.77e-5 \\
            & Batch Size       & \{32, 512\}       & 32          & 32          & 64             & 512 \\
            & Weight Decay     & \{1e-7, 1e-1\}    & 1.31e-2     & 4.43e-5     & 5.48e-4        & 1.36e-6 \\
            & Dropout          & \{0.0, 0.4\}      & 0.051       & 0.159       & 0.175          & 0.186 \\
            & Frozen Layers    & \{none, all\}     & N/A         & N/A         & None           & Last 5 layers \\
\bottomrule
\end{tabular}
\end{subtable}

\bigskip

\begin{subtable}{\textwidth}
\centering
\caption{Training hyperparameters for the sim-to-real \gadras{}$\to$experiment adaptation using a {\LaBr} detector.}
\label{table:sim2real_labr_hps}
\begin{tabular}{l l l l l l l}
\toprule
& & & \multicolumn{2}{c}{Best run (target-only)} & \multicolumn{2}{c}{Best run (domain-adapted)} \\
\cmidrule(lr){4-5} \cmidrule(lr){6-7}
Architecture & Parameter & Search Space & {\textit{N}=64} & {\textit{N}=1024} & {\textit{N}=64} & {\textit{N}=1024} \\
\midrule
MLP     & Learning Rate    & \{1e-6, 1e-3\}    & 1.19e-6     & 1.44e-6     & 8.10e-6     & 2.78e-4 \\
        & Batch Size       & \{32, 512\}       & 64          & 32          & 64          & 256 \\
        & Weight Decay     & \{1e-7, 1e-1\}    & 1.45e-6     & 8.26e-2     & 4.24e-7     & 3.80e-5 \\
        & Dropout          & \{0.0, 0.4\}      & 0.021       & 0.098       & 0.000312    & 0.344 \\
        & Frozen Layers    & \{none, all\}     & N/A         & N/A         & Last 2 layers & Last layer \\
\midrule
CNN     & Learning Rate    & \{1e-6, 1e-3\}    & 2.69e-6     & 5.06e-5     & 1.13e-4     & 4.49e-4 \\
        & Batch Size       & \{32, 512\}       & 64          & 32          & 32          & 256 \\
        & Weight Decay     & \{1e-7, 1e-1\}    & 9.62e-5     & 3.58e-2     & 1.06e-5     & 2.33e-5 \\
        & Dropout          & \{0.0, 0.4\}      & 0.000239    & 0.0349      & 0.00476     & 0.389 \\
        & Frozen Layers    & \{none, all\}     & N/A         & N/A         & First 3 layers & None \\
\midrule
TBNN (Li)& Learning Rate    & \{1e-6, 1e-3\}    & 2.65e-5     & 8.79e-4     & 5.07e-4     & 8.85e-4 \\
        & Batch Size       & \{32, 512\}       & 32          & 128         & 32          & 256 \\
        & Weight Decay     & \{1e-7, 1e-1\}    & 2.62e-4     & 6.98e-5     & 1.10e-2     & 1.30e-4 \\
        & Dropout          & \{0.0, 0.4\}      & 0.0117      & 0.0125      & 0.156       & 0.0125 \\
        & Frozen Layers    & \{none, all\}     & N/A         & N/A         & Last layer  & Last 8 layers \\
\midrule
TBNN (ours)    & Learning Rate    & \{1e-6, 1e-3\}    & 6.95e-6     & 2.25e-4     & 7.02e-4     & 2.37e-4 \\
        & Batch Size       & \{32, 512\}       & 32          & 32          & 32          & 128 \\
        & Weight Decay     & \{1e-7, 1e-1\}    & 1.63e-4     & 2.76e-5     & 4.54e-7     & 1.49e-4 \\
        & Dropout          & \{0.0, 0.4\}      & 0.000792    & 0.0347      & 0.0476      & 0.370 \\
        & Frozen Layers    & \{none, all\}     & N/A         & N/A         & Last 19 layers & Last layer \\
\bottomrule
\end{tabular}
\end{subtable}

\bigskip

\begin{subtable}{\textwidth}
\centering
\caption{Training hyperparameters for the sim-to-real \gadras{}$\to$experiment adaptation using a {\NaI} detector.}
\label{table:sim2real_nai_hps}
\begin{tabular}{l l l l l l l}
\toprule
& & & \multicolumn{2}{c}{Best run (target-only)} & \multicolumn{2}{c}{Best run (domain-adapted)} \\
\cmidrule(lr){4-5} \cmidrule(lr){6-7}
Architecture & Parameter & Search Space & {\textit{N}=64} & {\textit{N}=1024} & {\textit{N}=64} & {\textit{N}=1024} \\
\midrule
MLP     & Learning Rate    & \{1e-6, 1e-3\}    & 1.01e-6     & 7.79e-6     & 2.37e-4     & 4.84e-4 \\
        & Batch Size       & \{32, 512\}       & 32          & 32          & 64          & 32 \\
        & Weight Decay     & \{1e-7, 1e-1\}    & 4.05e-7     & 1.78e-2     & 1.49e-4     & 4.03e-4 \\
        & Dropout          & \{0.0, 0.4\}      & 0.012       & 0.019       & 0.370       & 0.141 \\
        & Frozen Layers    & \{none, all\}     & N/A         & N/A         & None        & Last layer \\
\midrule
CNN     & Learning Rate    & \{1e-6, 1e-3\}    & 2.14e-6     & 1.66e-4     & 7.43e-6     & 2.27e-4 \\
        & Batch Size       & \{32, 512\}       & 64          & 256         & 64          & 32 \\
        & Weight Decay     & \{1e-7, 1e-1\}    & 2.32e-3     & 4.70e-3     & 4.61e-5     & 4.51e-5 \\
        & Dropout          & \{0.0, 0.4\}      & 0.00335     & 0.00206     & 0.0120      & 0.00117 \\
        & Frozen Layers    & \{none, all\}     & N/A         & N/A         & Last layer  & None \\
\midrule
TBNN (Li)& Learning Rate    & \{1e-6, 1e-3\}    & 6.39e-4     & 3.75e-4     & 1.84e-4     & 7.74e-4 \\
        & Batch Size       & \{32, 512\}       & 64          & 64          & 64          & 32 \\
        & Weight Decay     & \{1e-7, 1e-1\}    & 1.30e-4     & 4.65e-4     & 2.33e-4     & 7.06e-4 \\
        & Dropout          & \{0.0, 0.4\}      & 0.00170     & 0.0298      & 0.00506     & 0.0383 \\
        & Frozen Layers    & \{none, all\}     & N/A         & N/A         & First 5 layers & First 13 layers \\
\midrule
TBNN (ours)    & Learning Rate    & \{1e-6, 1e-3\}    & 9.29e-5     & 2.35e-4     & 7.31e-4     & 2.53e-4 \\
        & Batch Size       & \{32, 512\}       & 64          & 32          & 32          & 64 \\
        & Weight Decay     & \{1e-7, 1e-1\}    & 2.13e-2     & 6.41e-7     & 4.98e-2     & 6.62e-5 \\
        & Dropout          & \{0.0, 0.4\}      & 0.00192     & 0.0164      & 0.184       & 0.350 \\
        & Frozen Layers    & \{none, all\}     & N/A         & N/A         & First 3 layers  & None \\
\bottomrule
\end{tabular}
\end{subtable}

\end{minipage}
}
\end{table}
\endgroup

\begin{figure}
    \centering
    \includegraphics[width=0.49\textwidth]{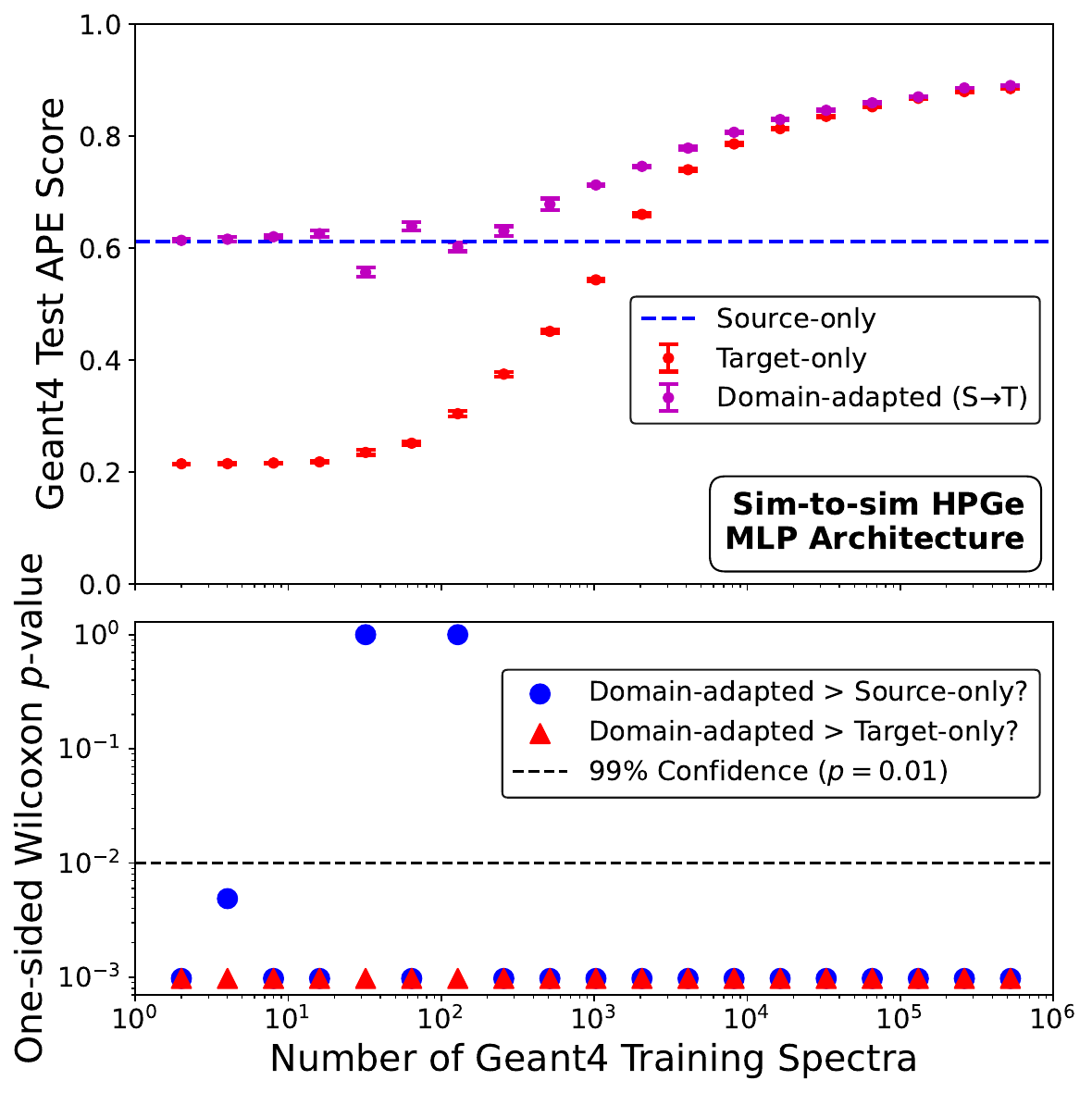}
    \hfill
    \includegraphics[width=0.49\textwidth]{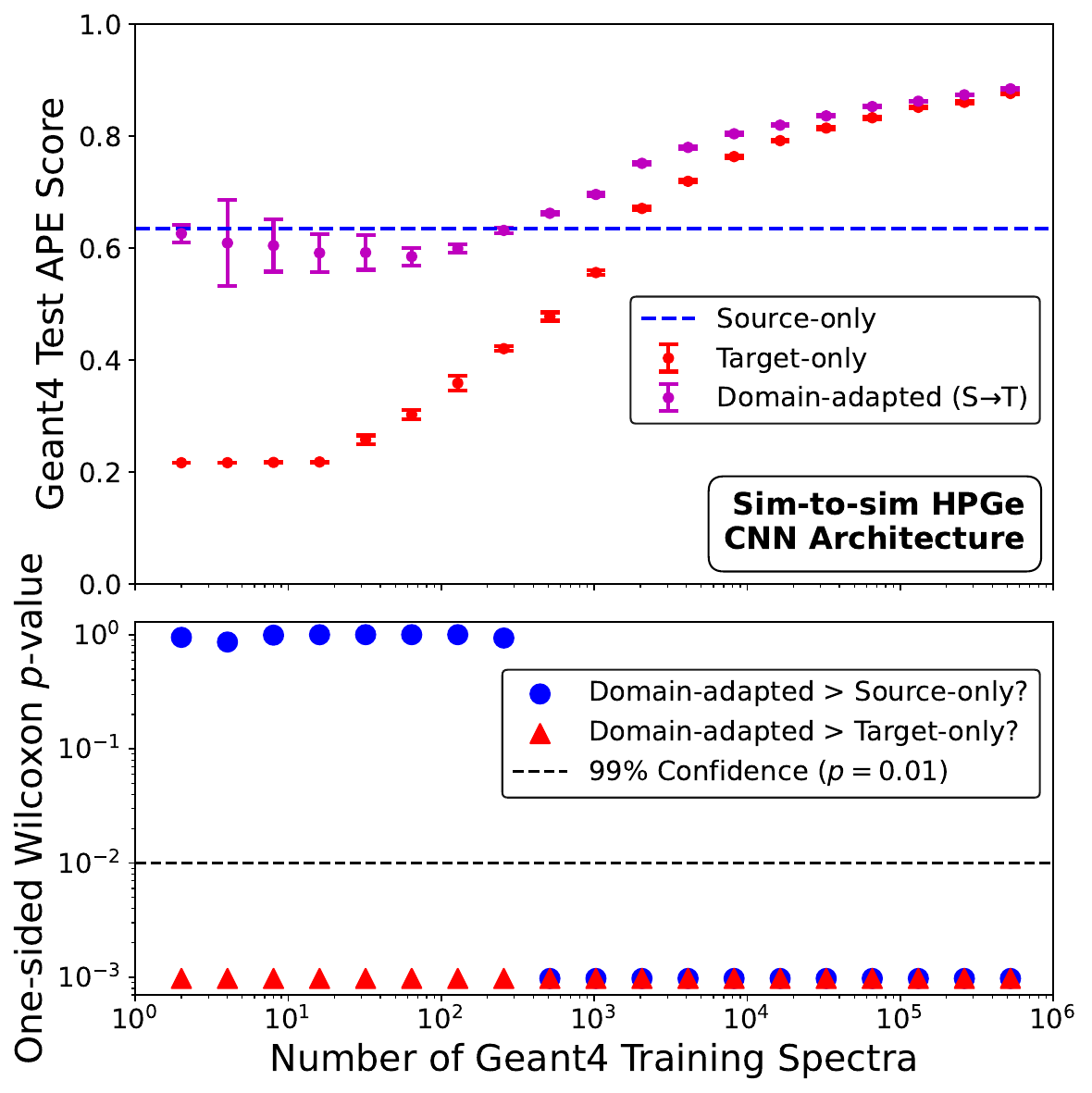}
    \vskip\baselineskip
    \includegraphics[width=0.49\textwidth]{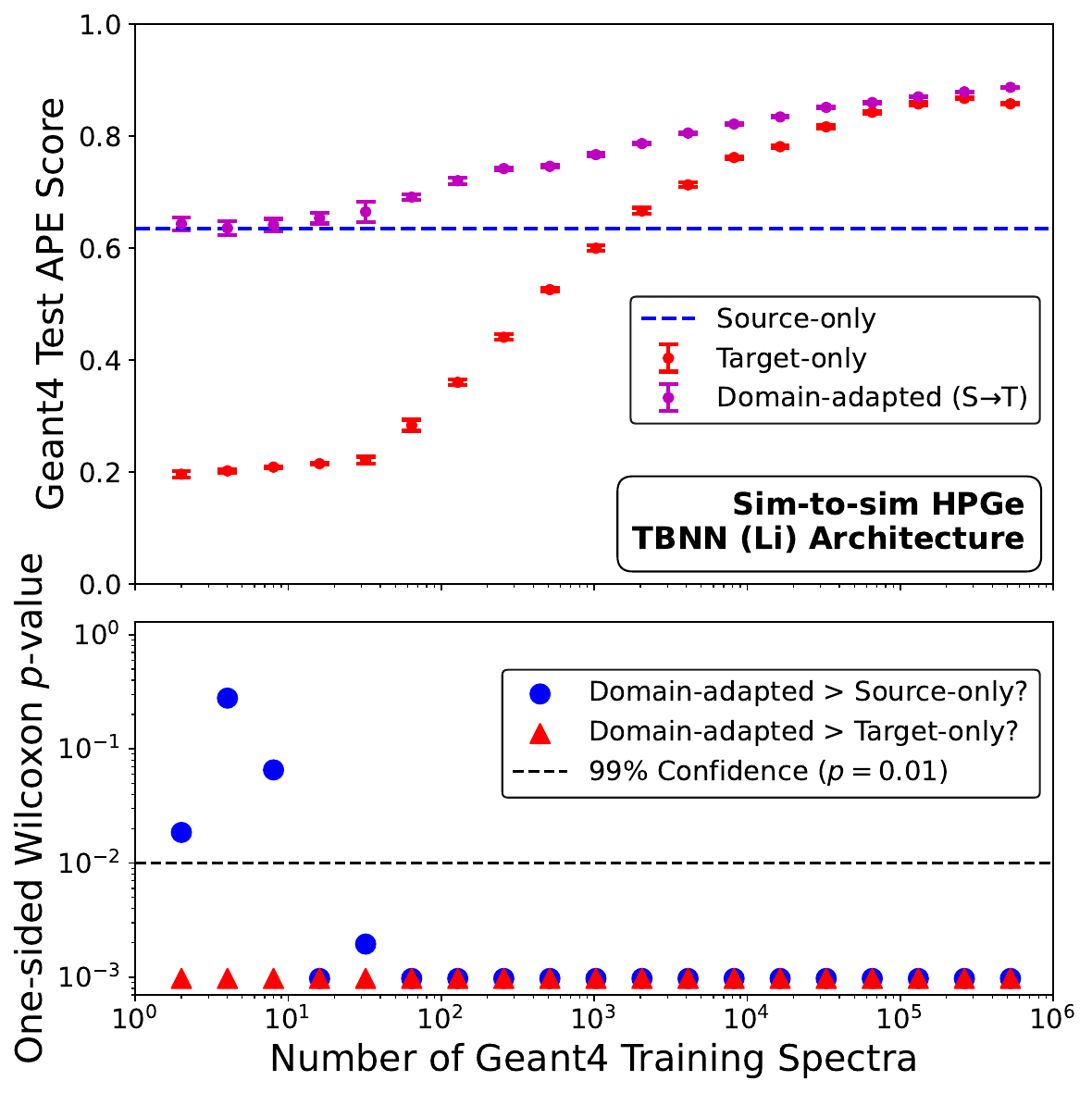}
    \hfill
    \includegraphics[width=0.49\textwidth]{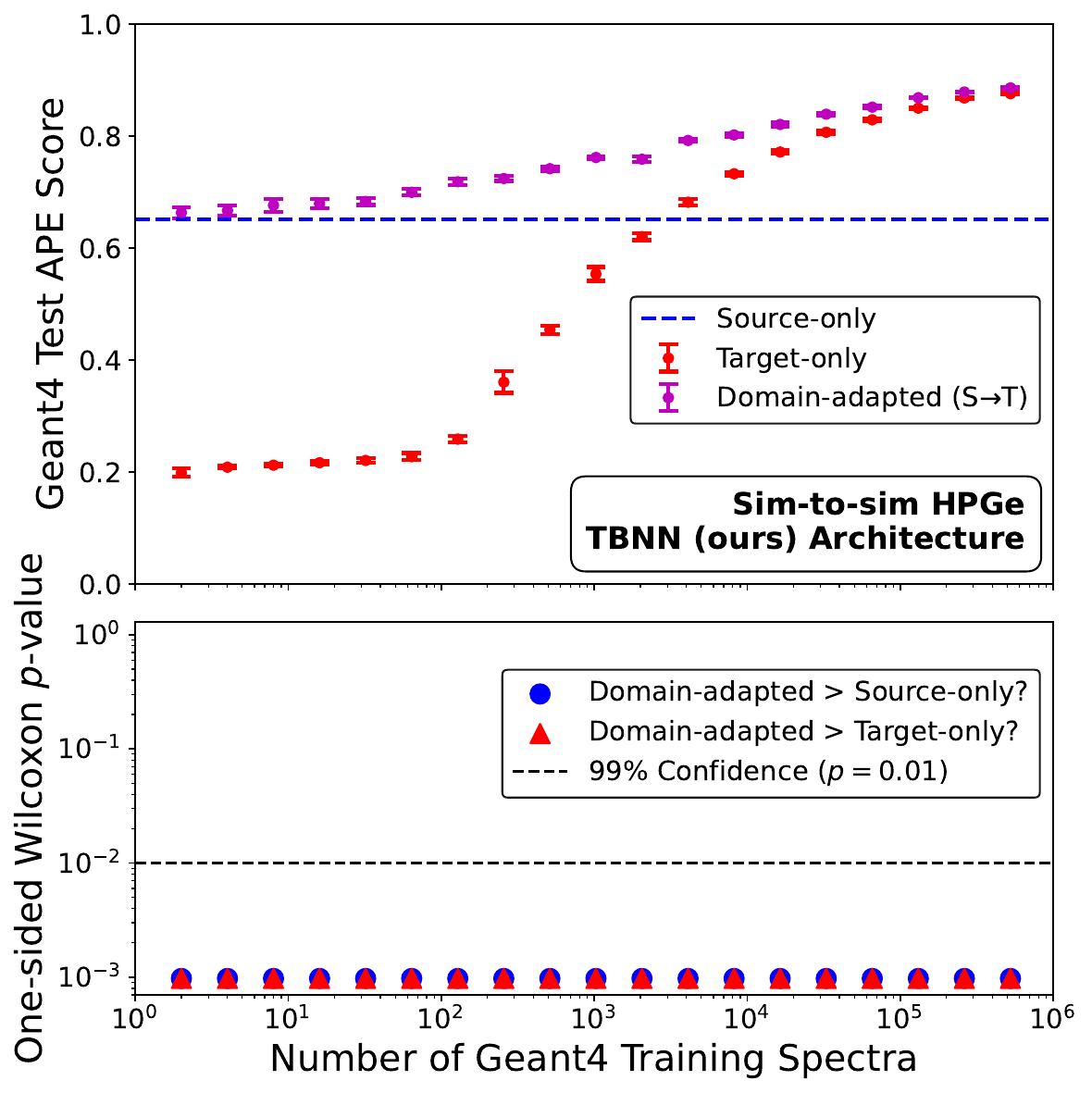}
    \caption{Performance comparison of domain-adapted versus baseline models for the \gadras{}$\to$\geant{} domain adaptation scenario. Top panels show the APE scores (Eq.~\ref{eq:APE_score}) on the \geant{} test dataset versus training size for three approaches: source-only models trained exclusively on \gadras{} data (dashed blue line), target-only models trained exclusively on \geant{} data (red error bars), and domain-adapted models using \gadras{} pretraining followed by \geant{} adaptation (magenta error bars). Error bars indicate the mean and sample standard deviation from 10 repeated trials with randomly drawn \geant{} training samples and different random weight initializations. Bottom panels display one-sided Wilcoxon signed-rank test $p$-values comparing the testing APE scores of domain-adapted models against source-only (blue circles) and target-only (red triangles) baselines, with the dashed horizontal line at $p = 0.01$ indicating $99\%$ confidence. Results are demonstrated across four neural network architectures (MLP, CNN, TBNN (Li), TBNN (ours)) showing that domain-adapted models consistently achieve equivalent or superior performance to both baseline training approaches.}
    \label{fig:combined_ape_hypothesis}
\end{figure}

\begin{figure}
    \centering
    \includegraphics[width=0.49\textwidth]{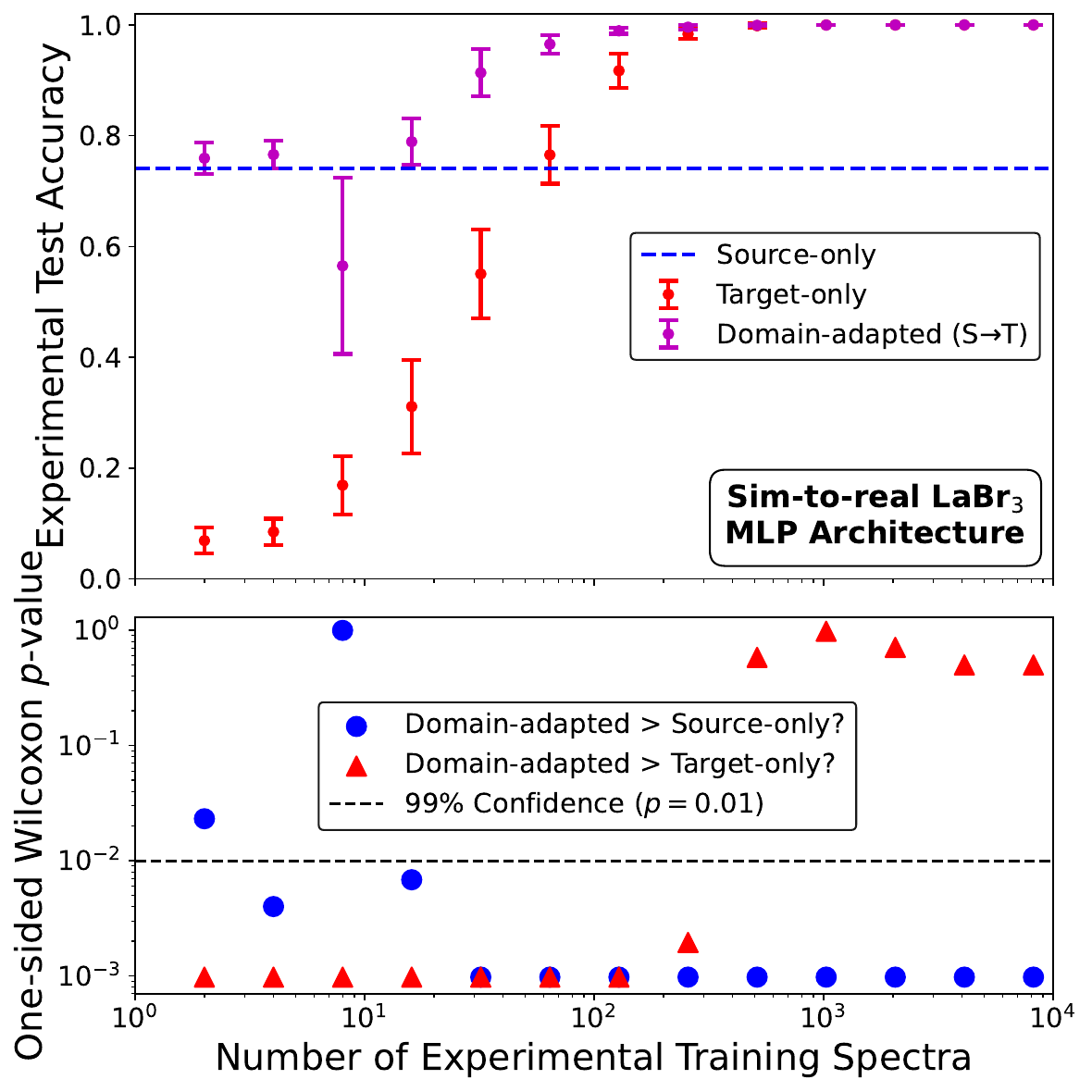}
    \hfill
    \includegraphics[width=0.49\textwidth]{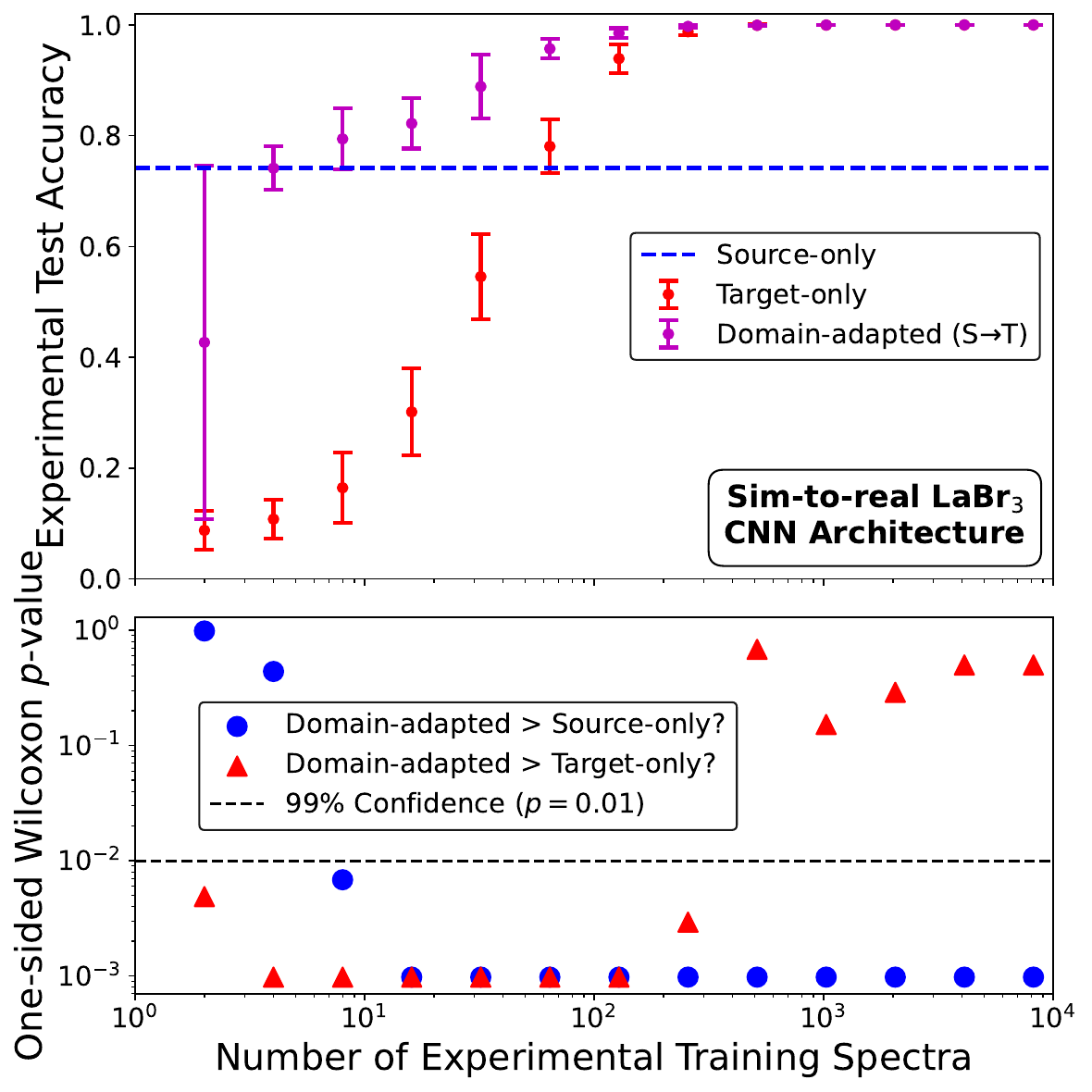}
    \vskip\baselineskip
    \includegraphics[width=0.49\textwidth]{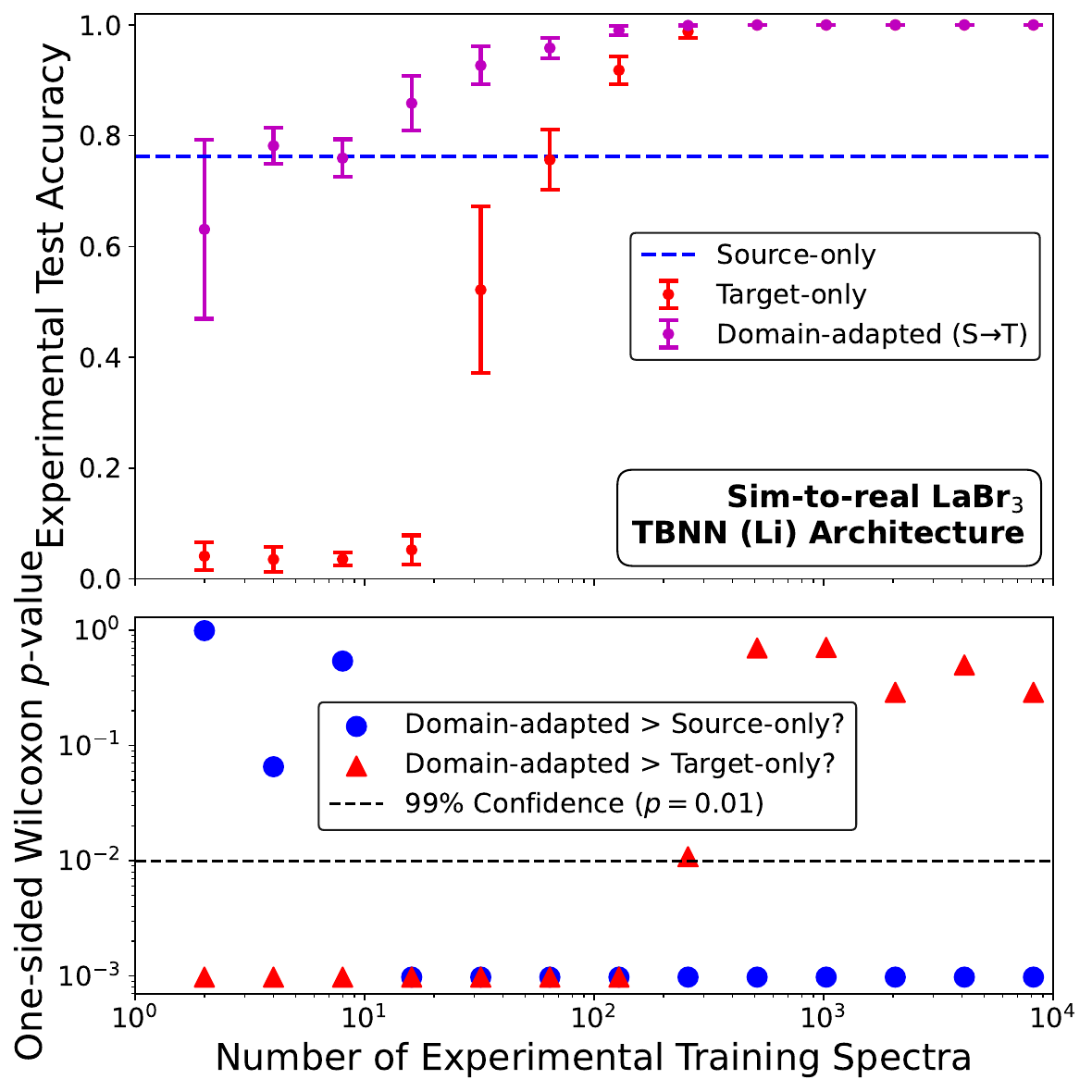}
    \hfill
    \includegraphics[width=0.49\textwidth]{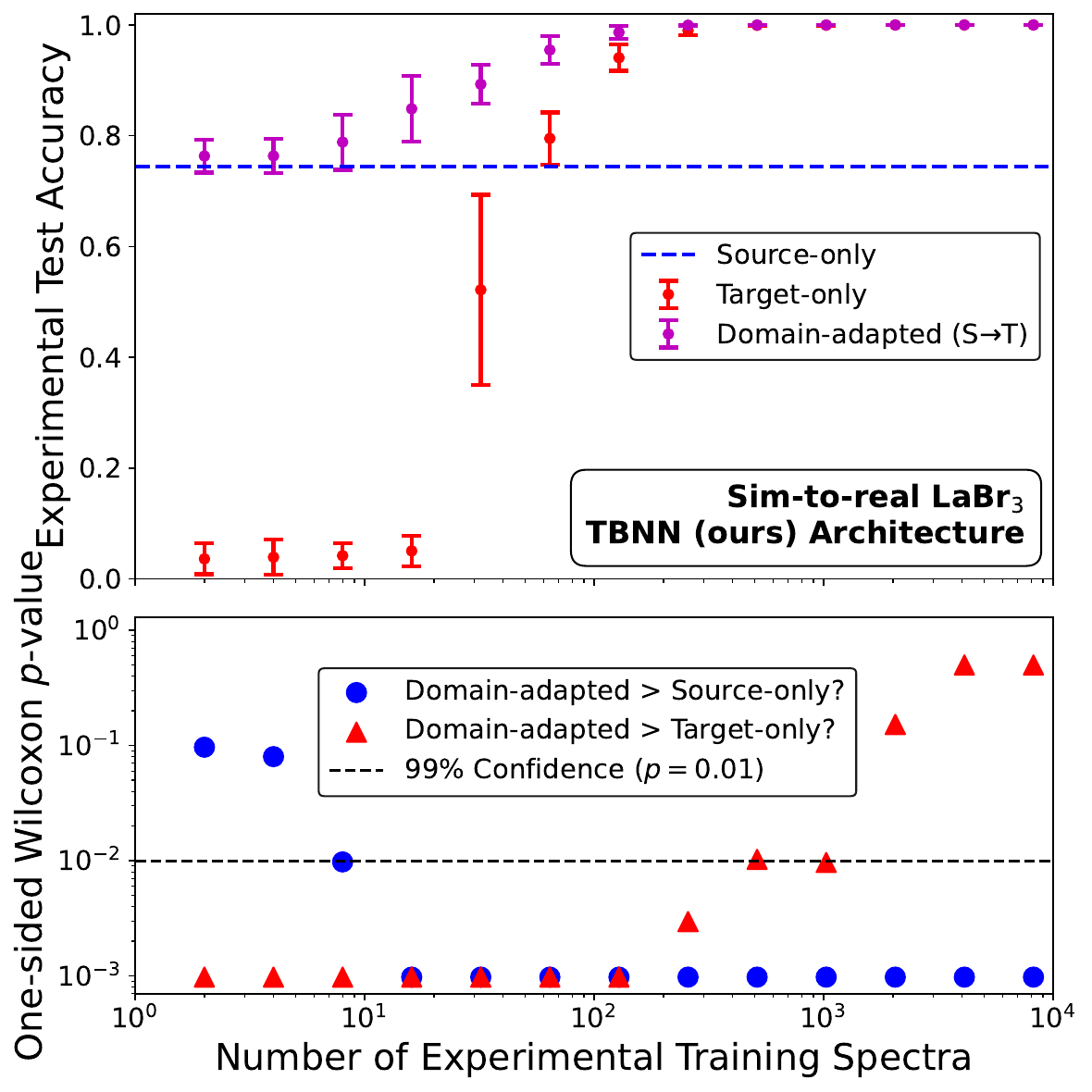}
    \caption{Same as Fig.~\ref{fig:combined_ape_hypothesis}, but for the \gadras{}$\to$experimental domain adaptation with a {\LaBr} detector. Performance is measured by experimental testing accuracy.}
    \label{fig:combined_accuracy_hypothesis_LaBr}
\end{figure}

\begin{figure}
    \centering
    \includegraphics[width=0.49\textwidth]{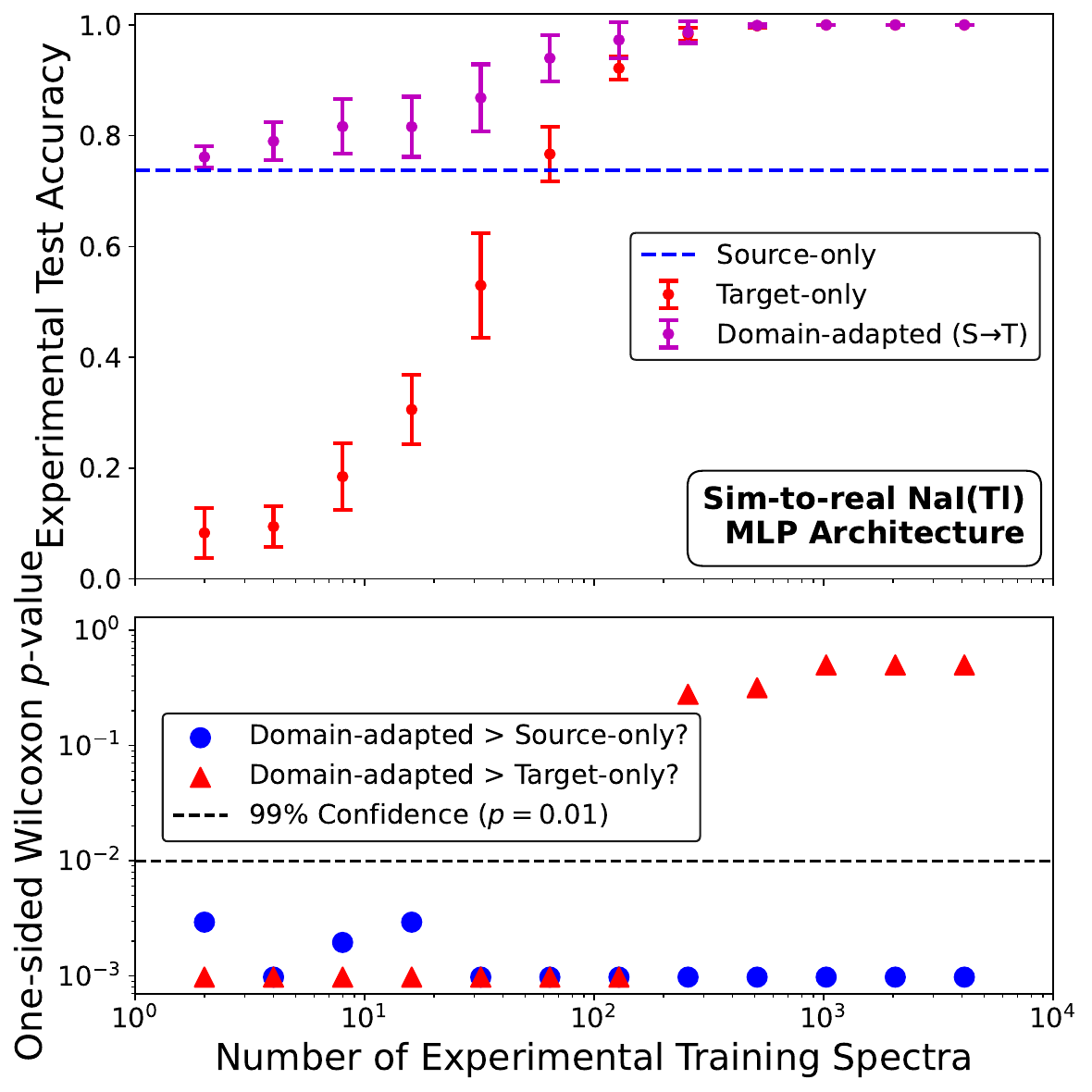}
    \hfill
    \includegraphics[width=0.49\textwidth]{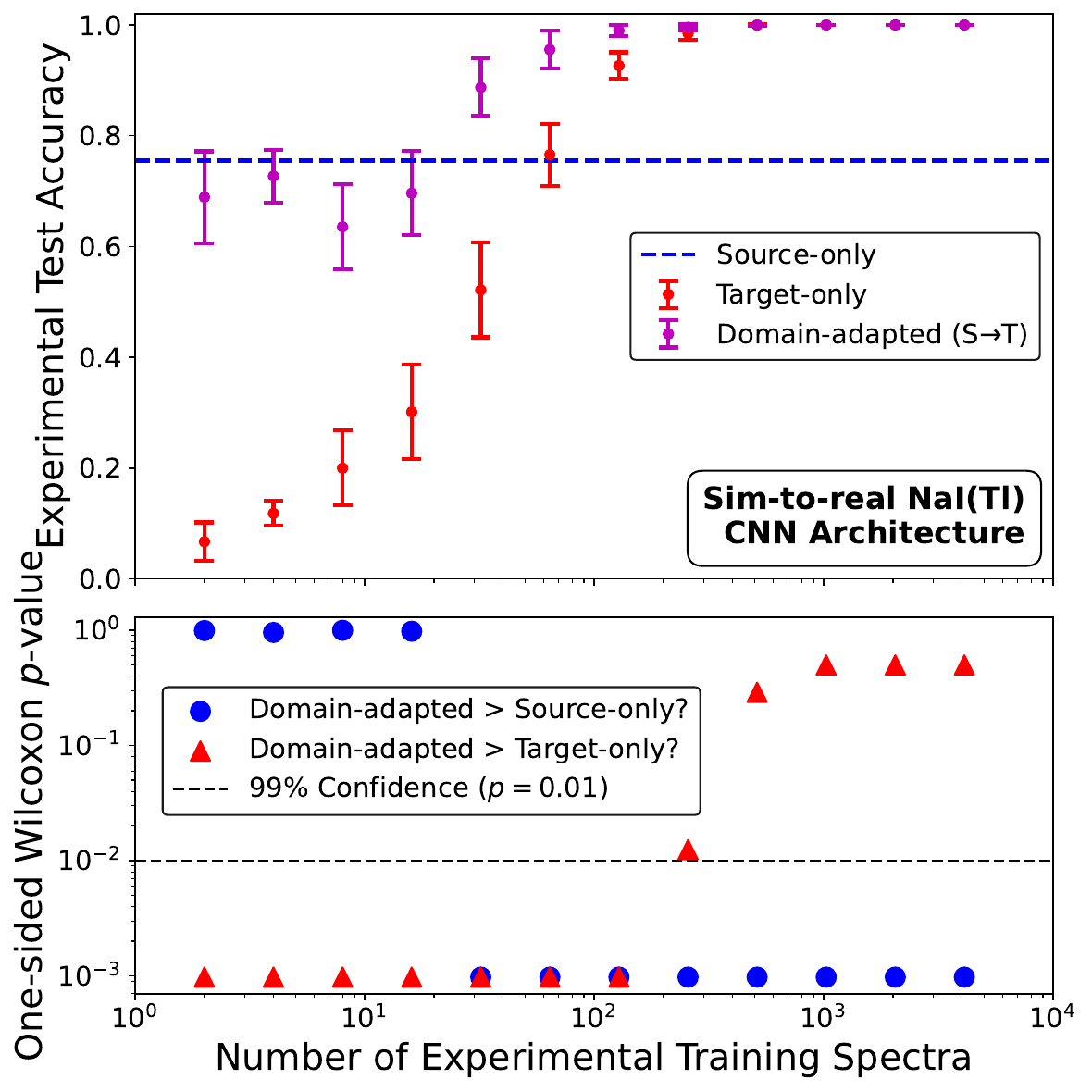}
    \vskip\baselineskip
    \includegraphics[width=0.49\textwidth]{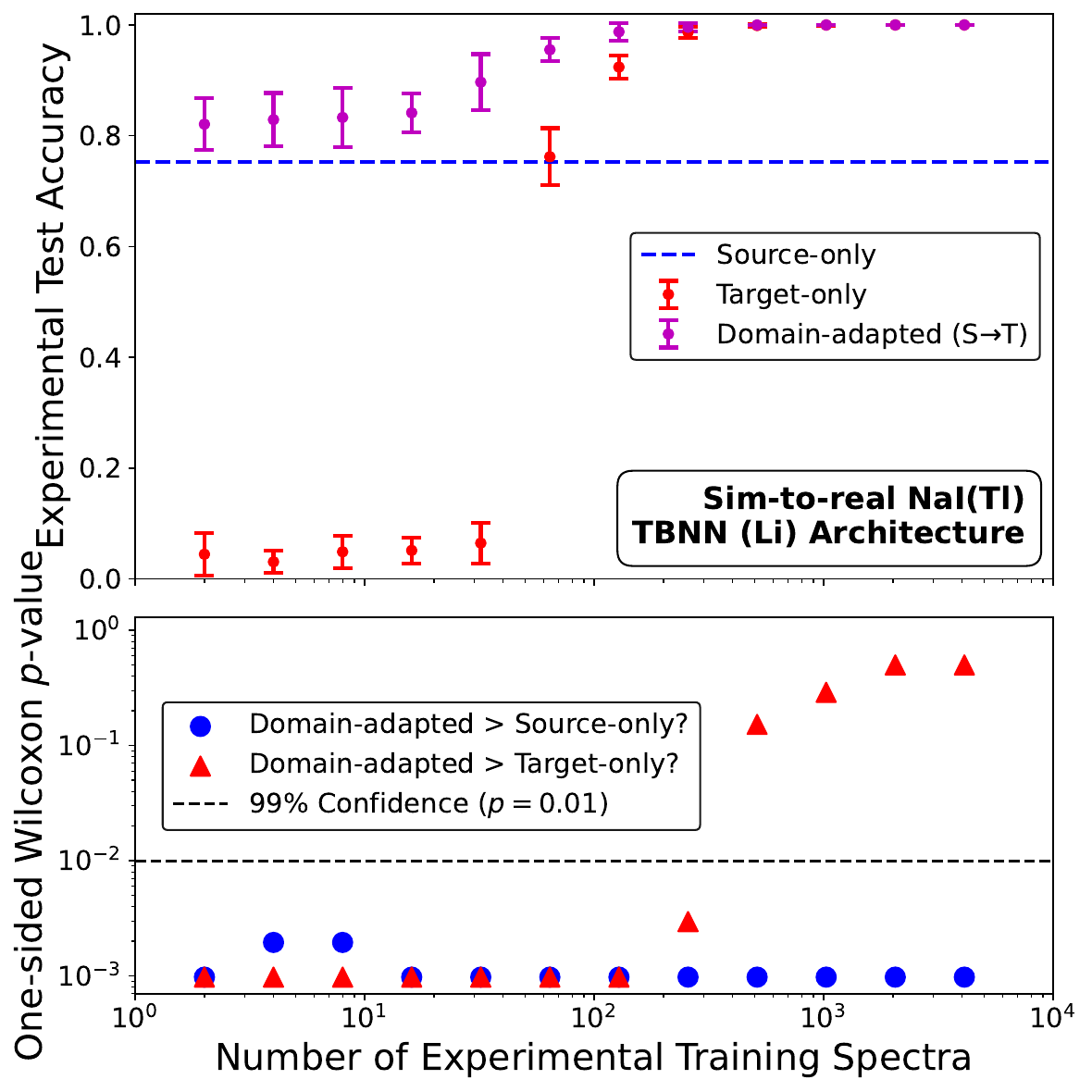}
    \hfill
    \includegraphics[width=0.49\textwidth]{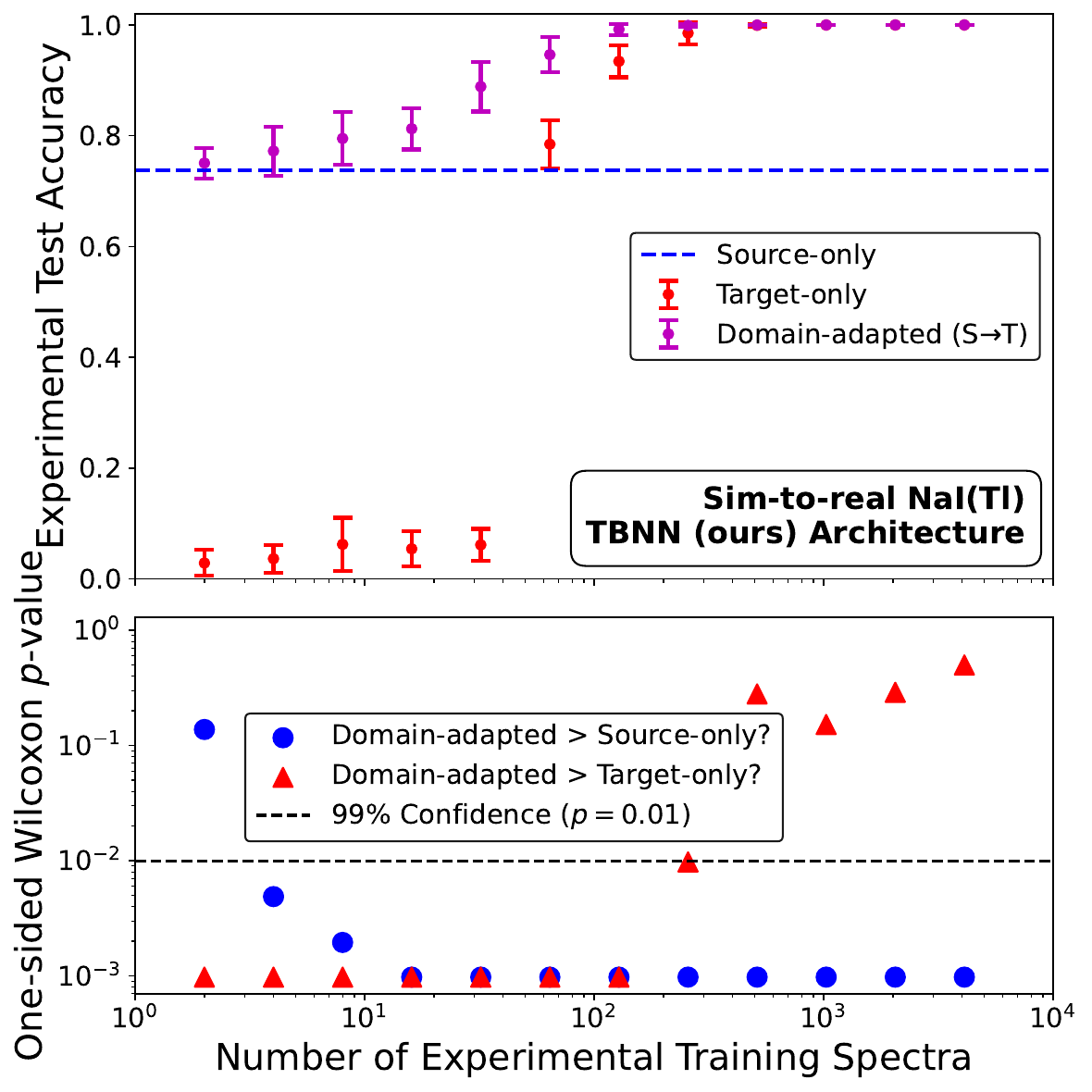}
    \caption{Same as Fig.~\ref{fig:combined_ape_hypothesis}, but for the \gadras{}$\to$experimental domain adaptation with a {\NaI} detector. Performance is measured by experimental testing accuracy.}
    \label{fig:combined_accuracy_hypothesis_NaI}
\end{figure}


\begin{figure}
    \centering
    
    \begin{subfigure}{0.6\textwidth} 
        \centering
        \includegraphics[width=\textwidth]{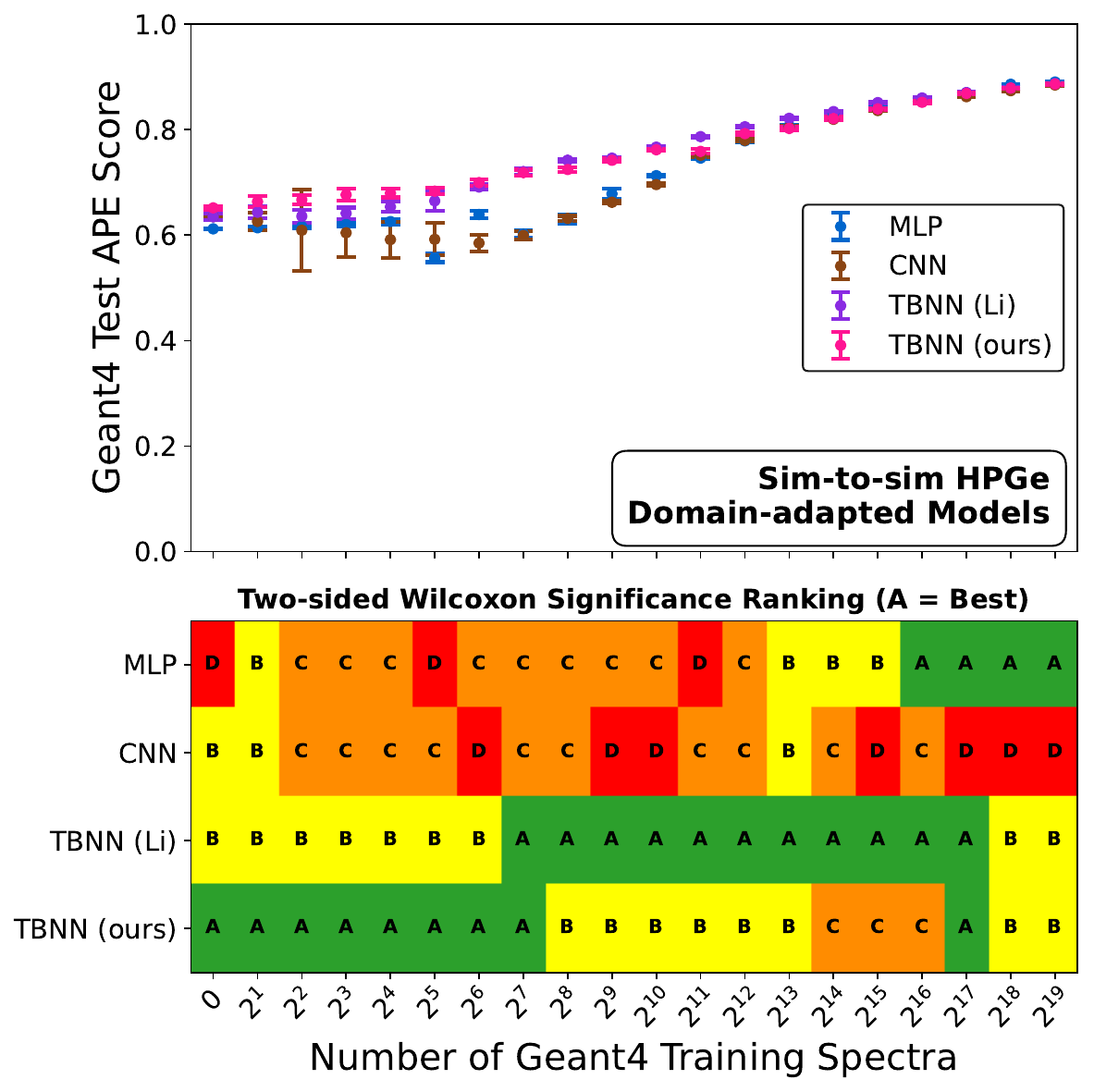}
        \caption{Architectural ranking for the sim-to-sim \gadras{}$\to$\geant{} adaptation using a {\HPGe} detector.}
        \label{fig:rank_geant}
    \end{subfigure}
    
    \vspace{0.5cm} 
    
    \begin{subfigure}{0.49\textwidth}
        \centering
        \includegraphics[width=\textwidth]{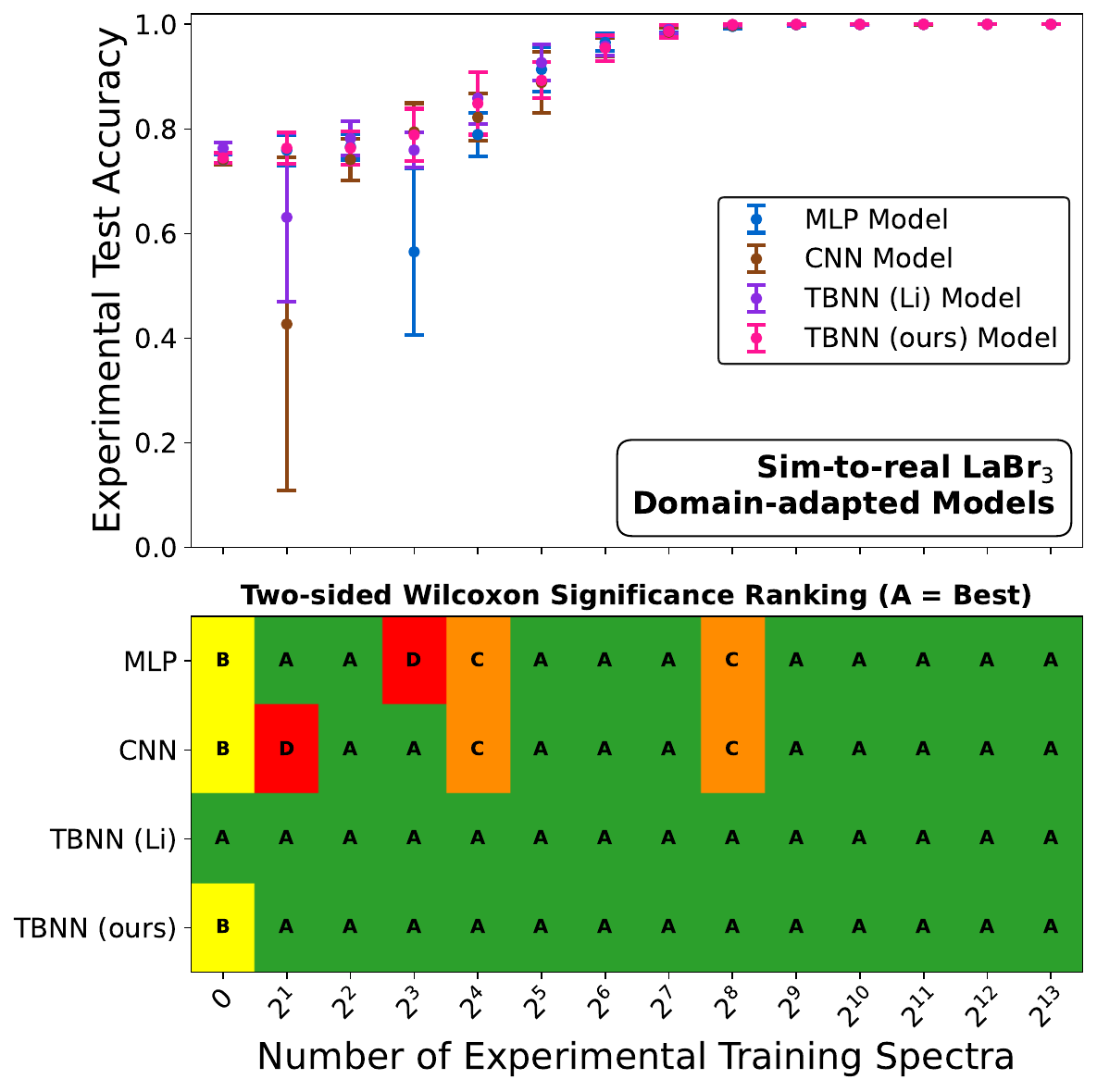}
        \caption{Architectural ranking for the sim-to-real \gadras{}$\to$experiment adaptation using a {\LaBr} detector.}
        \label{fig:rank_labr}
    \end{subfigure}
    \hfill 
    \begin{subfigure}{0.49\textwidth}
        \centering
        \includegraphics[width=\textwidth]{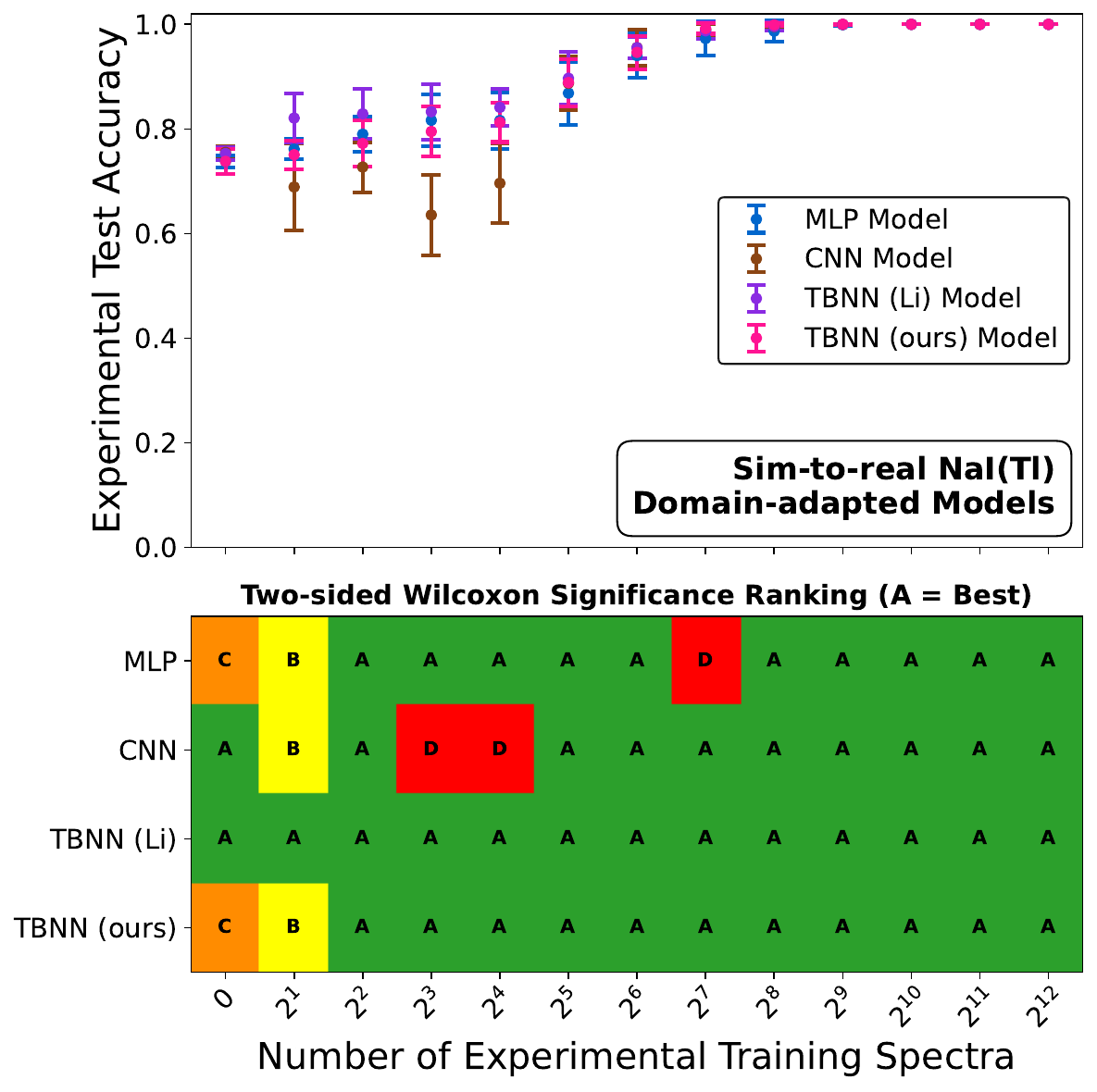}
        \caption{Architectural ranking for the sim-to-real \gadras{}$\to$experiment adaptation using a {\NaI} detector.}
        \label{fig:rank_nai}
    \end{subfigure}
    
    
    \caption{Statistical ranking of neural network architectures across training dataset sizes. Top panels show testing scores with error bars indicating the standard deviation across 10 random seeds. Bottom panels display statistical rankings using a two-sided Wilcoxon test $(\alpha = 0.01)$, where architectures sharing the same letter (A, B, C, D) are not significantly different. Results are shown for domain-adapted models for all domain adaptation scenarios: (a) sim-to-sim (\HPGe), (b) sim-to-real (\LaBr), and (c) sim-to-real (\NaI).}
    \label{fig:statistical_rank_combined}
\end{figure}

\begin{table}
\centering
\caption{Calculated $p$-values from a one-sided Wilcoxon signed-rank test, evaluating if domain-adapted models yield improved evaluation metric scores compared to the indicated baseline model (source-only or target-only). $p$-values $\le 0.01$ are bolded, indicating that the null hypothesis of no improvement can be rejected. Comparisons are shown for all domain adaptation scenarios: (a) sim-to-sim (\HPGe), (b) sim-to-real (\LaBr), and (c) sim-to-real (\NaI).}
\label{table:pvalue_ft_vs_baseline}

\begingroup
\linespread{1.1}\selectfont

\begin{subtable}{\textwidth}
\centering
\caption{$p$-values for the sim-to-sim \gadras{}$\to$\geant{} adaptation using a {\HPGe} detector.}
\adjustbox{max width=\linewidth}{
    \begin{tabular}{c c c c c c c c c}
    \toprule
    & \multicolumn{2}{c}{\textbf{MLP}} & \multicolumn{2}{c}{\textbf{CNN}} & \multicolumn{2}{c}{\textbf{TBNN (Li)}} & \multicolumn{2}{c}{\textbf{TBNN (ours)}} \\
    \shortstack{Target \\ Dataset Size} & \shortstack{Domain-adapted \\ vs. \\ source-only} & \shortstack{Domain-adapted \\ vs. \\ target-only} & \shortstack{Domain-adapted \\ vs. \\ source-only} & \shortstack{Domain-adapted \\ vs. \\ target-only} & \shortstack{Domain-adapted \\ vs. \\ source-only} & \shortstack{Domain-adapted \\ vs. \\ target-only} & \shortstack{Domain-adapted \\ vs. \\ source-only} & \shortstack{Domain-adapted \\ vs. \\ target-only} \\
    \midrule
    $2^1$ & $\mathbf{0.001}$ & $\mathbf{0.001}$ & $0.947$ & $\mathbf{0.001}$ & $0.019$ & $\mathbf{0.001}$ & $\mathbf{0.001}$ & $\mathbf{0.001}$ \\
    $2^2$ & $\mathbf{0.005}$ & $\mathbf{0.001}$ & $0.862$ & $\mathbf{0.001}$ & $0.278$ & $\mathbf{0.001}$ & $\mathbf{0.001}$ & $\mathbf{0.001}$ \\
    $2^3$ & $\mathbf{0.001}$ & $\mathbf{0.001}$ & $0.990$ & $\mathbf{0.001}$ & $0.065$ & $\mathbf{0.001}$ & $\mathbf{0.001}$ & $\mathbf{0.001}$ \\
    $2^4$ & $\mathbf{0.001}$ & $\mathbf{0.001}$ & $0.999$ & $\mathbf{0.001}$ & $\mathbf{0.001}$ & $\mathbf{0.001}$ & $\mathbf{0.001}$ & $\mathbf{0.001}$ \\
    $2^5$ & $1.000$ & $\mathbf{0.001}$ & $1.000$ & $\mathbf{0.001}$ & $\mathbf{0.002}$ & $\mathbf{0.001}$ & $\mathbf{0.001}$ & $\mathbf{0.001}$ \\
    $2^6$ & $\mathbf{0.001}$ & $\mathbf{0.001}$ & $1.000$ & $\mathbf{0.001}$ & $\mathbf{0.001}$ & $\mathbf{0.001}$ & $\mathbf{0.001}$ & $\mathbf{0.001}$ \\
    $2^7$ & $1.000$ & $\mathbf{0.001}$ & $1.000$ & $\mathbf{0.001}$ & $\mathbf{0.001}$ & $\mathbf{0.001}$ & $\mathbf{0.001}$ & $\mathbf{0.001}$ \\
    $2^8$ & $\mathbf{0.001}$ & $\mathbf{0.001}$ & $0.935$ & $\mathbf{0.001}$ & $\mathbf{0.001}$ & $\mathbf{0.001}$ & $\mathbf{0.001}$ & $\mathbf{0.001}$ \\
    $2^9$ to $2^{19}$ & $\mathbf{0.001}$ & $\mathbf{0.001}$ & $\mathbf{0.001}$ & $\mathbf{0.001}$ & $\mathbf{0.001}$ & $\mathbf{0.001}$ & $\mathbf{0.001}$ & $\mathbf{0.001}$ \\
    \bottomrule
    \end{tabular}
}
\end{subtable}

\bigskip

\begin{subtable}{\textwidth}
\centering
\caption{$p$-values for the sim-to-real \gadras{}$\to$experiment adaptation using a {\LaBr} detector.}
\adjustbox{max width=\linewidth}{
    \begin{tabular}{c c c c c c c c c}
    \toprule
    & \multicolumn{2}{c}{\textbf{MLP}} & \multicolumn{2}{c}{\textbf{CNN}} & \multicolumn{2}{c}{\textbf{TBNN (Li)}} & \multicolumn{2}{c}{\textbf{TBNN (ours)}} \\
    \shortstack{Target \\ Dataset Size} & \shortstack{Domain-adapted \\ vs. \\ source-only} & \shortstack{Domain-adapted \\ vs. \\ target-only} & \shortstack{Domain-adapted \\ vs. \\ source-only} & \shortstack{Domain-adapted \\ vs. \\ target-only} & \shortstack{Domain-adapted \\ vs. \\ source-only} & \shortstack{Domain-adapted \\ vs. \\ target-only} & \shortstack{Domain-adapted \\ vs. \\ source-only} & \shortstack{Domain-adapted \\ vs. \\ target-only} \\
    \midrule
    $2^1$ & $0.023$ & $\mathbf{0.001}$ & $0.986$ & $\mathbf{0.005}$ & $0.993$ & $\mathbf{0.001}$ & $0.097$ & $\mathbf{0.001}$ \\
    $2^2$ & $\mathbf{0.004}$ & $\mathbf{0.001}$ & $0.439$ & $\mathbf{0.001}$ & $0.065$ & $\mathbf{0.001}$ & $0.080$ & $\mathbf{0.001}$ \\
    $2^3$ & $0.998$ & $\mathbf{0.001}$ & $\mathbf{0.007}$ & $\mathbf{0.001}$ & $0.541$ & $\mathbf{0.001}$ & $\mathbf{0.010}$ & $\mathbf{0.001}$ \\
    $2^4$ & $\mathbf{0.007}$ & $\mathbf{0.001}$ & $\mathbf{0.001}$ & $\mathbf{0.001}$ & $\mathbf{0.001}$ & $\mathbf{0.001}$ & $\mathbf{0.001}$ & $\mathbf{0.001}$ \\
    $2^5$ & $\mathbf{0.001}$ & $\mathbf{0.001}$ & $\mathbf{0.001}$ & $\mathbf{0.001}$ & $\mathbf{0.001}$ & $\mathbf{0.001}$ & $\mathbf{0.001}$ & $\mathbf{0.001}$ \\
    $2^6$ & $\mathbf{0.001}$ & $\mathbf{0.001}$ & $\mathbf{0.001}$ & $\mathbf{0.001}$ & $\mathbf{0.001}$ & $\mathbf{0.001}$ & $\mathbf{0.001}$ & $\mathbf{0.001}$ \\
    $2^7$ & $\mathbf{0.001}$ & $\mathbf{0.001}$ & $\mathbf{0.001}$ & $\mathbf{0.001}$ & $\mathbf{0.001}$ & $\mathbf{0.001}$ & $\mathbf{0.001}$ & $\mathbf{0.001}$ \\
    $2^8$ & $\mathbf{0.001}$ & $\mathbf{0.002}$ & $\mathbf{0.001}$ & $\mathbf{0.003}$ & $\mathbf{0.001}$ & $0.011$ & $\mathbf{0.001}$ & $\mathbf{0.003}$ \\
    $2^9$ & $\mathbf{0.001}$ & $0.582$ & $\mathbf{0.001}$ & $0.683$ & $\mathbf{0.001}$ & $0.702$ & $\mathbf{0.001}$ & $\mathbf{0.010}$ \\
    $2^{10}$ & $\mathbf{0.001}$ & $0.982$ & $\mathbf{0.001}$ & $0.152$ & $\mathbf{0.001}$ & $0.710$ & $\mathbf{0.001}$ & $\mathbf{0.010}$ \\
    $2^{11}$ & $\mathbf{0.001}$ & $0.710$ & $\mathbf{0.001}$ & $0.290$ & $\mathbf{0.001}$ & $0.290$ & $\mathbf{0.001}$ & $0.152$ \\
    $2^{12}$ & $\mathbf{0.001}$ & $0.500$ & $\mathbf{0.001}$ & $0.500$ & $\mathbf{0.001}$ & $0.500$ & $\mathbf{0.001}$ & $0.500$ \\
    $2^{13}$ & $\mathbf{0.001}$ & $0.500$ & $\mathbf{0.001}$ & $0.500$ & $\mathbf{0.001}$ & $0.290$ & $\mathbf{0.001}$ & $0.500$ \\
    \bottomrule
    \end{tabular}
}
\end{subtable}

\bigskip

\begin{subtable}{\textwidth}
\centering
\caption{$p$-values for the sim-to-real \gadras{}$\to$experiment adaptation using a {\NaI} detector.}
\adjustbox{max width=\linewidth}{
    \begin{tabular}{c c c c c c c c c}
    \toprule
    & \multicolumn{2}{c}{\textbf{MLP}} & \multicolumn{2}{c}{\textbf{CNN}} & \multicolumn{2}{c}{\textbf{TBNN (Li)}} & \multicolumn{2}{c}{\textbf{TBNN (ours)}} \\
    \shortstack{Target \\ Dataset Size} & \shortstack{Domain-adapted \\ vs. \\ source-only} & \shortstack{Domain-adapted \\ vs. \\ target-only} & \shortstack{Domain-adapted \\ vs. \\ source-only} & \shortstack{Domain-adapted \\ vs. \\ target-only} & \shortstack{Domain-adapted \\ vs. \\ source-only} & \shortstack{Domain-adapted \\ vs. \\ target-only} & \shortstack{Domain-adapted \\ vs. \\ source-only} & \shortstack{Domain-adapted \\ vs. \\ target-only} \\
    \midrule
    $2^1$ & $\mathbf{0.003}$ & $\mathbf{0.001}$ & $0.995$ & $\mathbf{0.001}$ & $\mathbf{0.001}$ & $\mathbf{0.001}$ & $0.138$ & $\mathbf{0.001}$ \\
    $2^2$ & $\mathbf{0.001}$ & $\mathbf{0.001}$ & $0.958$ & $\mathbf{0.001}$ & $\mathbf{0.002}$ & $\mathbf{0.001}$ & $\mathbf{0.005}$ & $\mathbf{0.001}$ \\
    $2^3$ & $\mathbf{0.002}$ & $\mathbf{0.001}$ & $1.000$ & $\mathbf{0.001}$ & $\mathbf{0.002}$ & $\mathbf{0.001}$ & $\mathbf{0.002}$ & $\mathbf{0.001}$ \\
    $2^4$ & $\mathbf{0.003}$ & $\mathbf{0.001}$ & $0.981$ & $\mathbf{0.001}$ & $\mathbf{0.001}$ & $\mathbf{0.001}$ & $\mathbf{0.001}$ & $\mathbf{0.001}$ \\
    $2^5$ & $\mathbf{0.001}$ & $\mathbf{0.001}$ & $\mathbf{0.001}$ & $\mathbf{0.001}$ & $\mathbf{0.001}$ & $\mathbf{0.001}$ & $\mathbf{0.001}$ & $\mathbf{0.001}$ \\
    $2^6$ & $\mathbf{0.001}$ & $\mathbf{0.001}$ & $\mathbf{0.001}$ & $\mathbf{0.001}$ & $\mathbf{0.001}$ & $\mathbf{0.001}$ & $\mathbf{0.001}$ & $\mathbf{0.001}$ \\
    $2^7$ & $\mathbf{0.001}$ & $\mathbf{0.001}$ & $\mathbf{0.001}$ & $\mathbf{0.001}$ & $\mathbf{0.001}$ & $\mathbf{0.001}$ & $\mathbf{0.001}$ & $\mathbf{0.001}$ \\
    $2^8$ & $\mathbf{0.001}$ & $0.278$ & $\mathbf{0.001}$ & $0.012$ & $\mathbf{0.001}$ & $\mathbf{0.003}$ & $\mathbf{0.001}$ & $\mathbf{0.010}$ \\
    $2^9$ & $\mathbf{0.001}$ & $0.317$ & $\mathbf{0.001}$ & $0.290$ & $\mathbf{0.001}$ & $0.152$ & $\mathbf{0.001}$ & $0.280$ \\
    $2^{10}$ & $\mathbf{0.001}$ & $0.500$ & $\mathbf{0.001}$ & $0.500$ & $\mathbf{0.001}$ & $0.290$ & $\mathbf{0.001}$ & $0.152$ \\
    $2^{11}$ & $\mathbf{0.001}$ & $0.500$ & $\mathbf{0.001}$ & $0.500$ & $\mathbf{0.001}$ & $0.500$ & $\mathbf{0.001}$ & $0.290$ \\
    $2^{12}$ & $\mathbf{0.001}$ & $0.500$ & $\mathbf{0.001}$ & $0.500$ & $\mathbf{0.001}$ & $0.500$ & $\mathbf{0.001}$ & $0.500$ \\
    \bottomrule
    \end{tabular}
}
\end{subtable}

\endgroup
\end{table}

\begin{table}
\centering
\caption{Calculated $p$-values from a two-sided Wilcoxon signed-rank test, comparing the architectural performance of the domain-adapted models. For each fine-tuning dataset size, the model with the best evaluation metric score is indicated as `BEST', and each other cell evaluates if the given architecture is statistically equivalent to BEST. $p$-values $\leq 0.01$ are bolded, identifying architectures that yield statistically worse testing scores than BEST. Comparisons are shown for all three domain adaptation scenarios: (a) sim-to-sim (\HPGe), (b) sim-to-real (\LaBr), and (c) sim-to-real (\NaI).}
\label{table:pvalue_arch_comparison}
\begingroup
\linespread{1.1}\selectfont
\begin{subtable}{\textwidth}
\centering
\caption{$p$-values for the sim-to-sim \gadras{}$\to$\geant{} domain adaptation using a {\HPGe} detector.}
\label{table:pvalue_arch_comparison_hpge}
\begin{tabular}{ccccc}
\toprule
\shortstack{Fine-tuning \\ Dataset \\ Size} & \shortstack{MLP \\ vs. \\ Best} & \shortstack{CNN \\ vs. \\ Best} & \shortstack{TBNN (Li) \\ vs. \\ Best} & \shortstack{TBNN (ours) \\ vs. \\ Best} \\
\midrule
0            & \textbf{0.002} & \textbf{0.002} & \textbf{0.002} & BEST \\
$2^{1}$      & \textbf{0.002} & \textbf{0.002} & \textbf{0.002} & BEST \\
$2^{2}$      & \textbf{0.002} & \textbf{0.002} & \textbf{0.002} & BEST \\
$2^{3}$      & \textbf{0.002} & \textbf{0.002} & \textbf{0.002} & BEST \\
$2^{4}$      & \textbf{0.002} & \textbf{0.002} & \textbf{0.002} & BEST \\
$2^{5}$      & \textbf{0.002} & \textbf{0.002} & \textbf{0.004} & BEST \\
$2^{6}$      & \textbf{0.002} & \textbf{0.002} & \textbf{0.002} & BEST \\
$2^{7}$      & \textbf{0.002} & \textbf{0.002} & BEST           & 0.492 \\
$2^{8}$      & \textbf{0.002} & \textbf{0.002} & BEST           & \textbf{0.002} \\
$2^{9}$      & \textbf{0.002} & \textbf{0.002} & BEST           & \textbf{0.010} \\
$2^{10}$     & \textbf{0.002} & \textbf{0.002} & BEST           & \textbf{0.002} \\
$2^{11}$     & \textbf{0.002} & \textbf{0.002} & BEST           & \textbf{0.002} \\
$2^{12}$     & \textbf{0.002} & \textbf{0.002} & BEST           & \textbf{0.002} \\
$2^{13}$     & \textbf{0.002} & \textbf{0.002} & BEST           & \textbf{0.002} \\
$2^{14}$     & \textbf{0.002} & \textbf{0.002} & BEST           & \textbf{0.002} \\
$2^{15}$     & \textbf{0.002} & \textbf{0.002} & BEST           & \textbf{0.002} \\
$2^{16}$     & 0.232          & \textbf{0.002} & BEST           & \textbf{0.002} \\
$2^{17}$     & BEST           & \textbf{0.002} & BEST           & 0.037 \\
$2^{18}$     & BEST           & \textbf{0.002} & \textbf{0.002} & \textbf{0.002} \\
$2^{19}$     & BEST           & \textbf{0.002} & \textbf{0.002} & \textbf{0.002} \\
\bottomrule
\end{tabular}
\end{subtable}

\bigskip

\begin{subtable}{0.49\textwidth}
\centering
\caption{$p$-values for the sim-to-real \gadras{}$\to$experiment domain adaptation using a {\LaBr} detector.}
\label{table:pvalue_arch_comparison_labr}
\adjustbox{max width=\linewidth}{
\begin{tabular}{@{}ccccc@{}}
\toprule
\shortstack{Fine-tuning \\ Dataset \\ Size} & \shortstack{MLP \\ vs. \\ Best} & \shortstack{CNN \\ vs. \\ Best} & \shortstack{TBNN (Li) \\ vs. \\ Best} & \shortstack{TBNN (ours) \\ vs. \\ Best} \\
\midrule
0        & \textbf{0.002} & \textbf{0.002} & BEST           & \textbf{0.002} \\
$2^{1}$  & BEST           & 0.027          & 0.084          & BEST \\
$2^{2}$  & 0.232          & \textbf{0.020} & BEST           & 0.084 \\
$2^{3}$  & \textbf{0.002} & BEST           & 0.027          & 0.625 \\
$2^{4}$  & \textbf{0.002} & \textbf{0.004} & BEST           & 0.415 \\
$2^{5}$  & 0.432          & 0.049          & BEST           & \textbf{0.010} \\
$2^{6}$  & BEST           & 0.027          & 0.221          & \textbf{0.006} \\
$2^{7}$  & BEST           & 0.041          & BEST           & 0.838 \\
$2^{8}$  & \textbf{0.011} & \textbf{0.014} & 0.441          & BEST \\
$2^{9}$  & 0.076          & \textbf{0.011} & 0.153          & BEST \\
$2^{10}$ & \textbf{0.036} & 0.074          & 0.305          & BEST \\
$2^{11}$ & 0.579          & 0.579          & BEST           & BEST \\
$2^{12}$ & BEST           & BEST           & BEST           & BEST \\
$2^{13}$ & BEST           & BEST           & BEST           & BEST \\
\bottomrule
\end{tabular}
}
\end{subtable}
\hfill
\begin{subtable}{0.49\textwidth}
\caption{$p$-values for the sim-to-real \gadras{}$\to$experiment domain adaptation using a {\NaI} detector.}
\label{table:pvalue_arch_comparison_nai}
\adjustbox{max width=\linewidth}{
\begin{tabular}{@{}ccccc@{}}
\toprule
\shortstack{Fine-tuning \\ Dataset \\ Size} & \shortstack{MLP \\ vs. \\ Best} & \shortstack{CNN \\ vs. \\ Best} & \shortstack{TBNN (Li) \\ vs. \\ Best} & \shortstack{TBNN (ours) \\ vs. \\ Best} \\
\midrule
0        & \textbf{0.002} & BEST           & 0.492          & \textbf{0.010} \\
$2^{1}$  & \textbf{0.002} & \textbf{0.002} & BEST           & \textbf{0.006} \\
$2^{2}$  & 0.027          & \textbf{0.002} & BEST           & 0.064 \\
$2^{3}$  & 0.275          & \textbf{0.002} & BEST           & 0.160 \\
$2^{4}$  & 0.105          & \textbf{0.002} & BEST           & \textbf{0.027} \\
$2^{5}$  & 0.193          & 0.432          & BEST           & 0.432 \\
$2^{6}$  & \textbf{0.010} & BEST           & 0.695          & 0.160 \\
$2^{7}$  & \textbf{0.002} & 0.153          & 0.281          & BEST \\
$2^{8}$  & \textbf{0.006} & \textbf{0.021} & 0.560          & BEST \\
$2^{9}$  & 0.305          & BEST           & BEST           & 0.579 \\
$2^{10}$ & BEST           & BEST           & BEST           & BEST \\
$2^{11}$ & BEST           & BEST           & BEST           & BEST \\
$2^{12}$ & BEST           & BEST           & BEST           & BEST \\
\bottomrule
\end{tabular}
}
\end{subtable}
\endgroup
\end{table}

\begin{figure}
    \centering
    \includegraphics[width=0.48\textwidth]{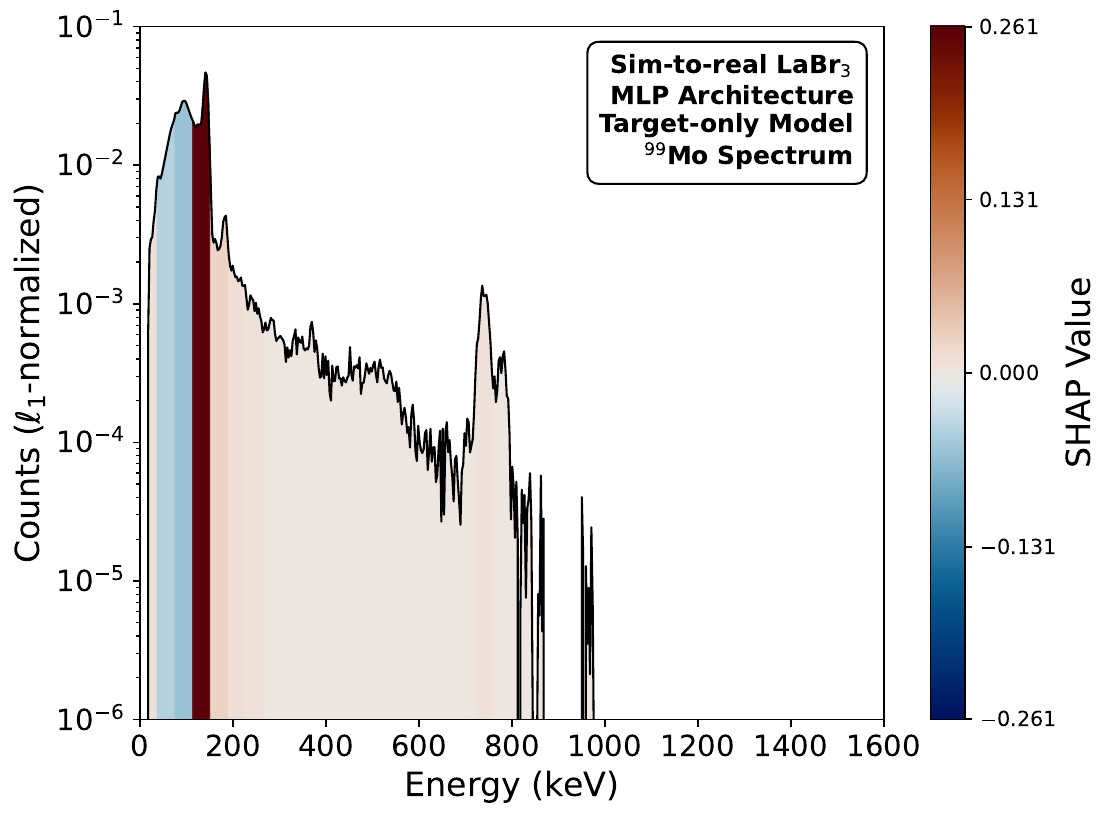}
    \includegraphics[width=0.48\textwidth]{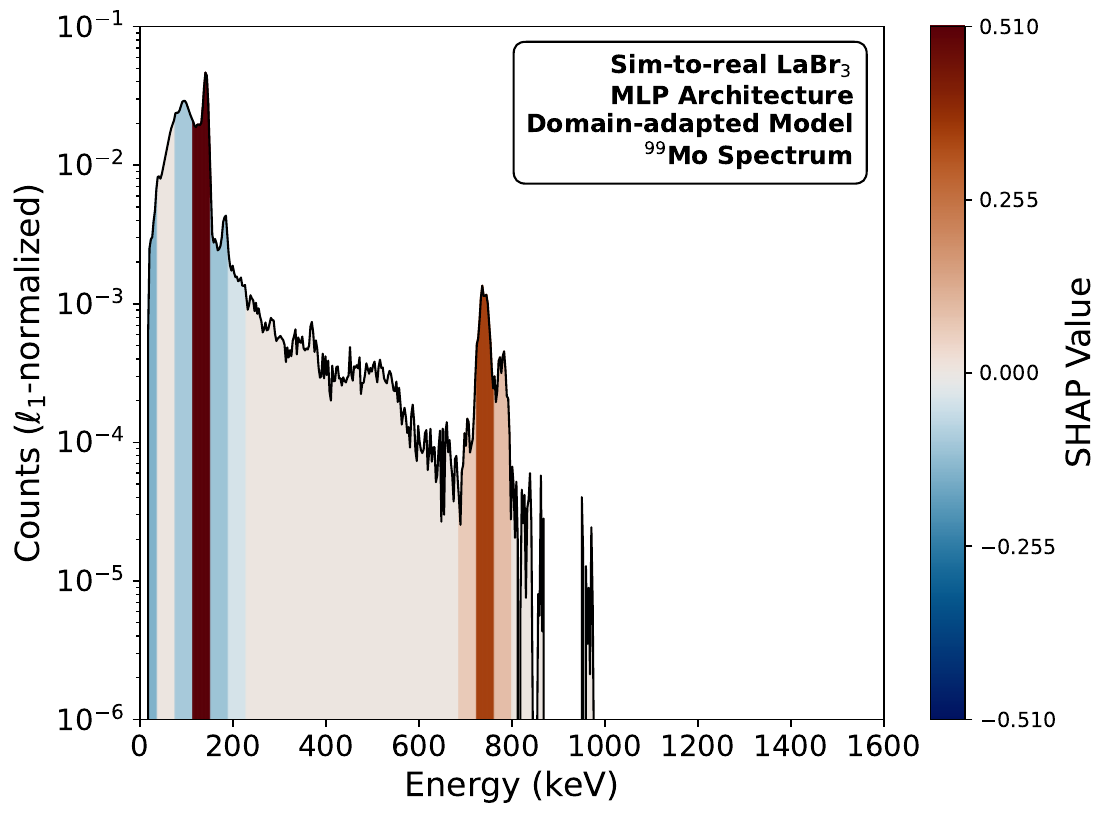}
    \includegraphics[width=0.48\textwidth]{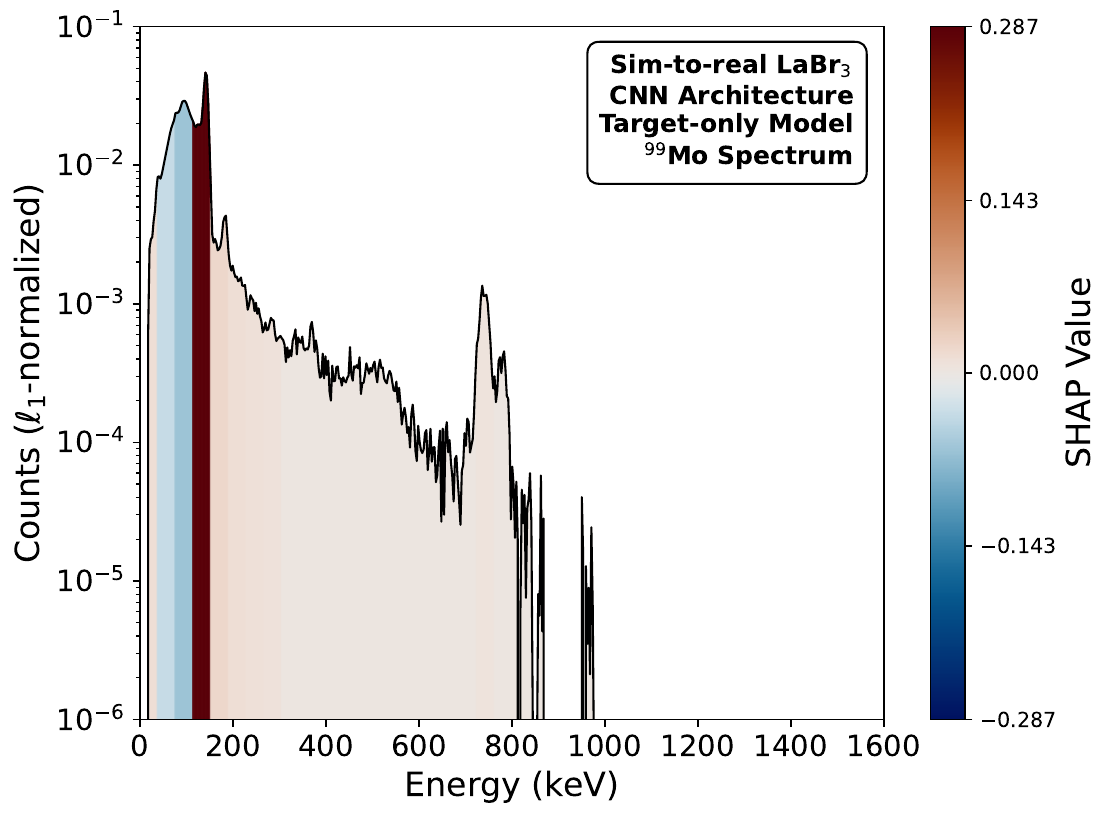}
    \includegraphics[width=0.48\textwidth]{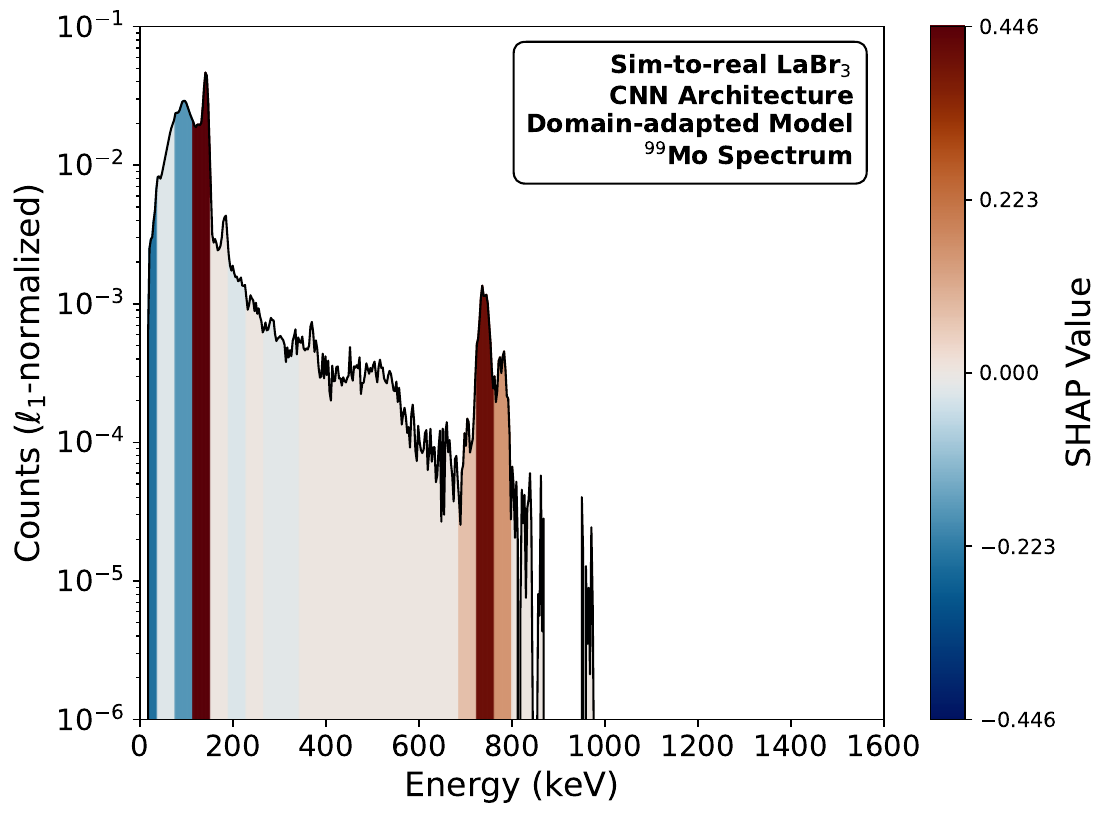}
    \includegraphics[width=0.48\textwidth]{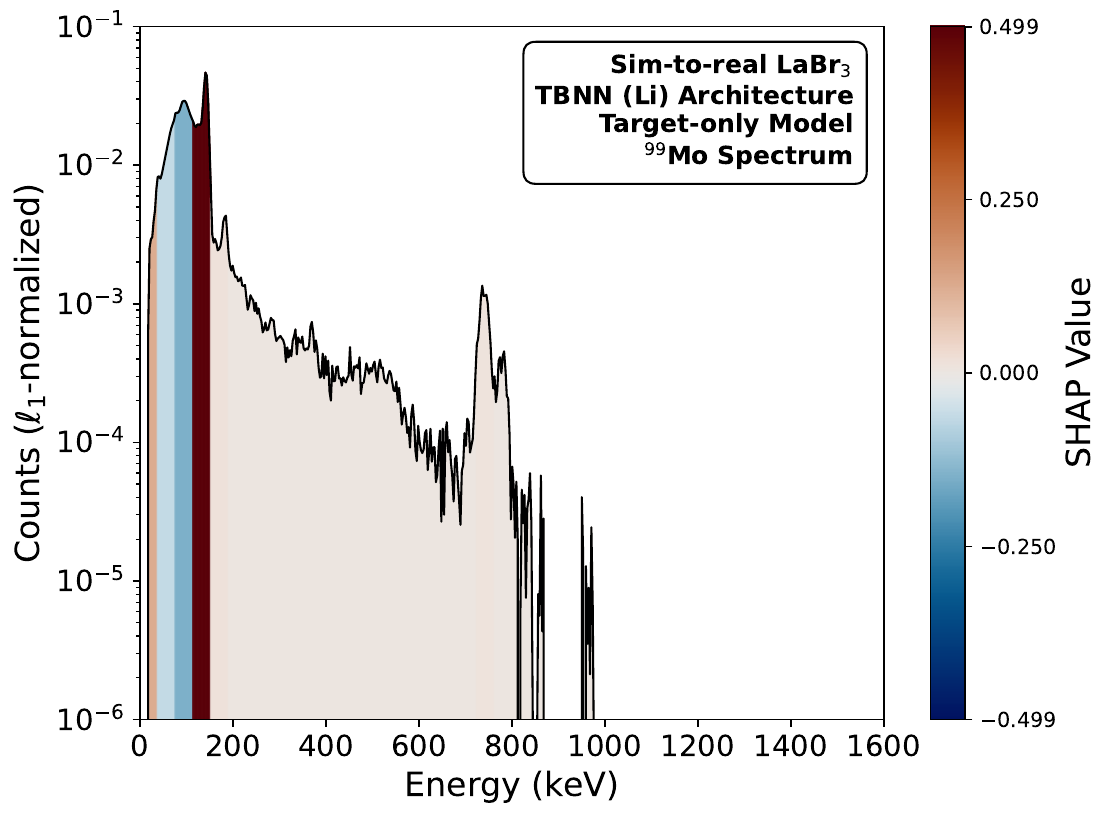}
    \includegraphics[width=0.48\textwidth]{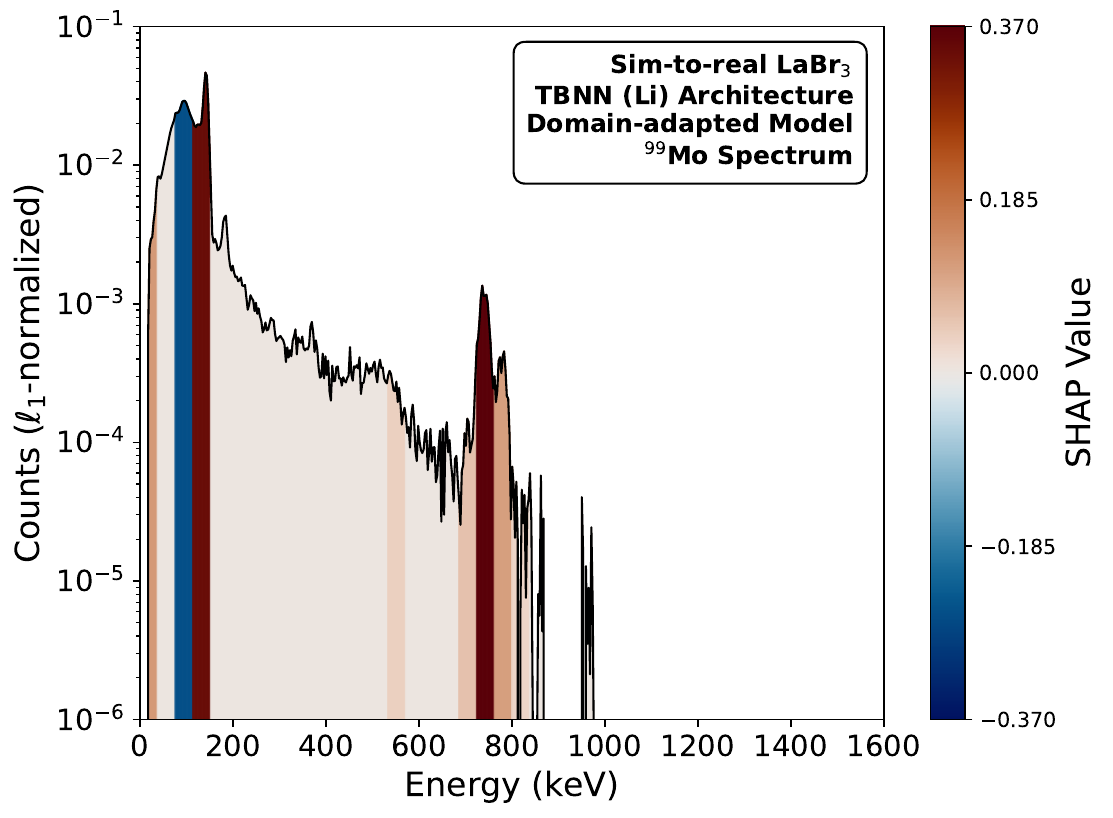}
    \caption{Same as Fig.~\ref{fig:XAI_Mo99}, but for an MLP architecture (top row), CNN architecture (middle row), and TBNN (Li) architecture (bottom row).}
    \label{fig:XAI_Mo99_all}
\end{figure}

\begin{figure}
    \centering
    \includegraphics[width=0.48\textwidth]{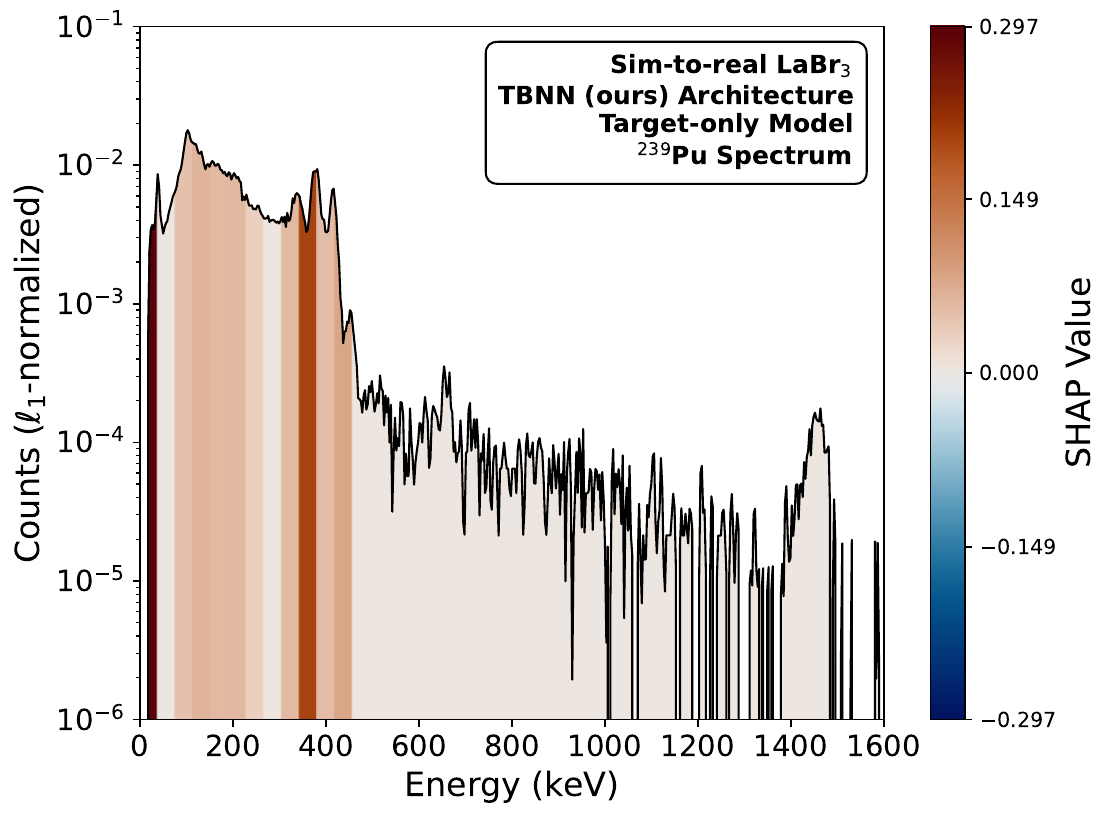}
    \includegraphics[width=0.48\textwidth]{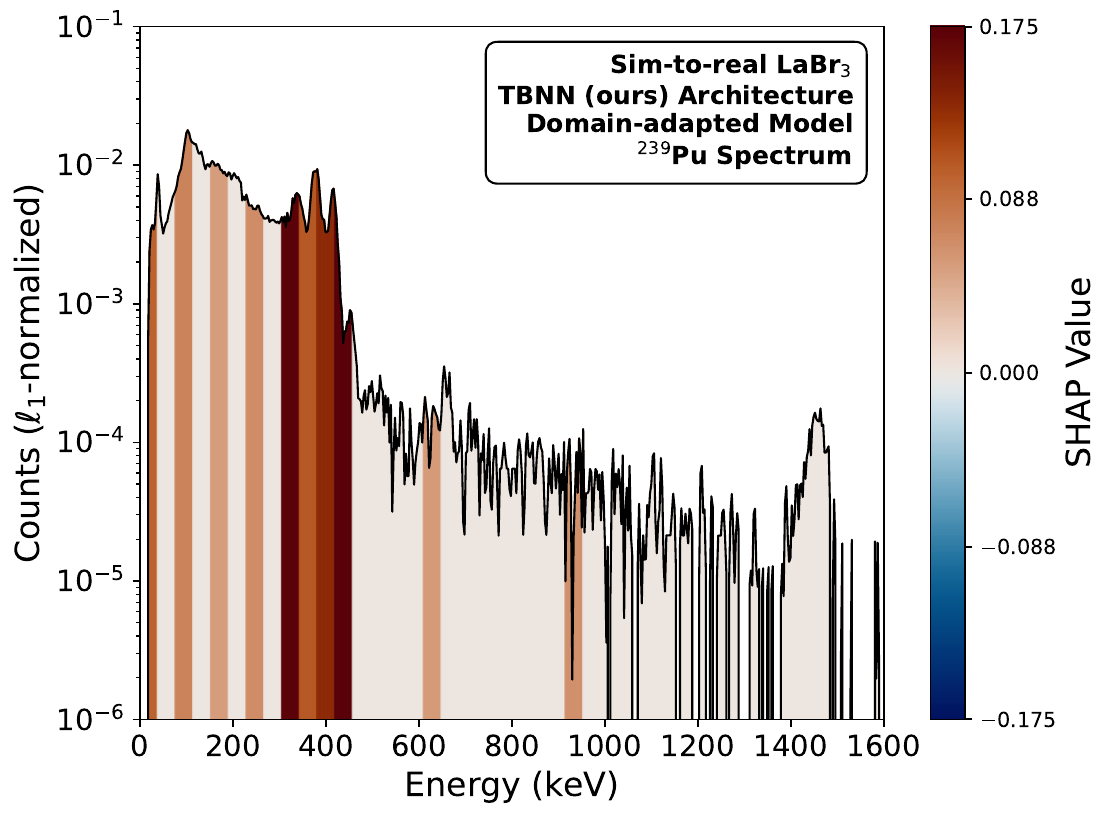}
    \caption{SHAP explanations for a $^{239}$Pu spectrum using a target-only model (left) and a domain-adapted model (right). The target-only model strongly weights on the $20$-$40$ keV band, whereas the domain-adapted model clearly highlights the 345, 375, and 413 keV peaks characteristic of $^{239}$Pu.}
    \label{fig:XAI_Pu239}
\end{figure}

\begin{figure}
    \centering
    \includegraphics[width=0.48\textwidth]{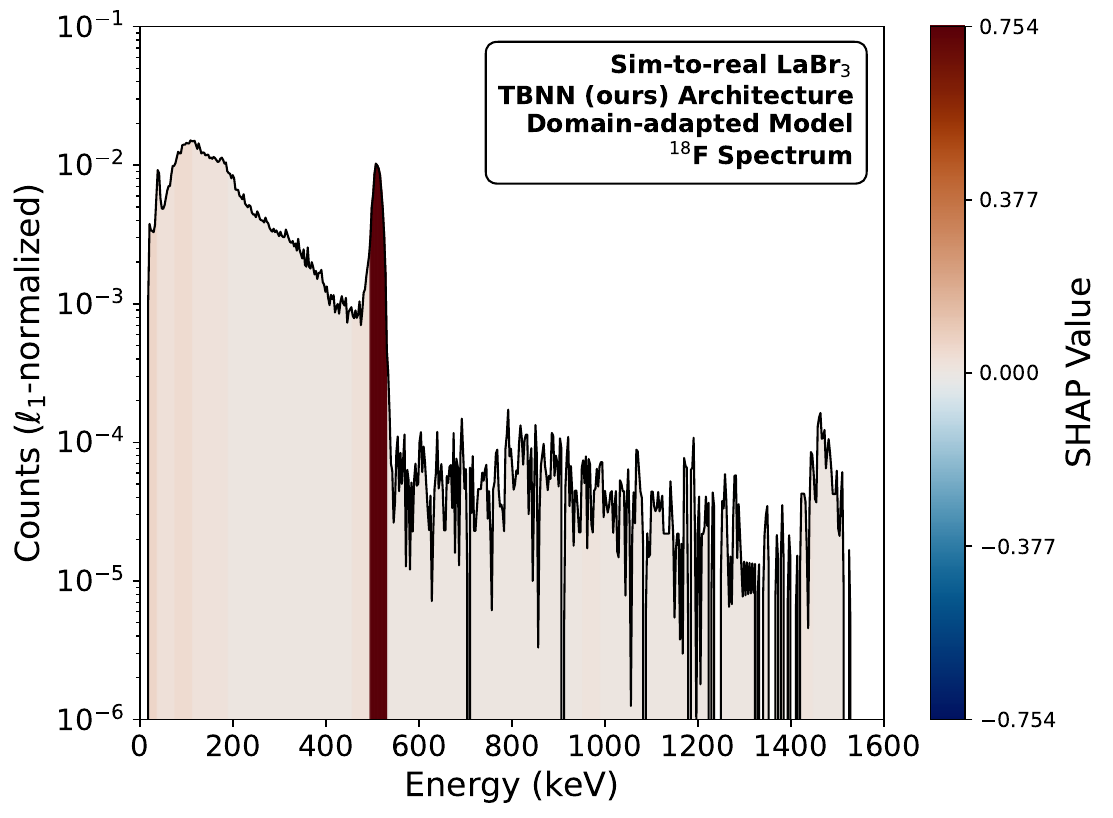}
    \includegraphics[width=0.48\textwidth]{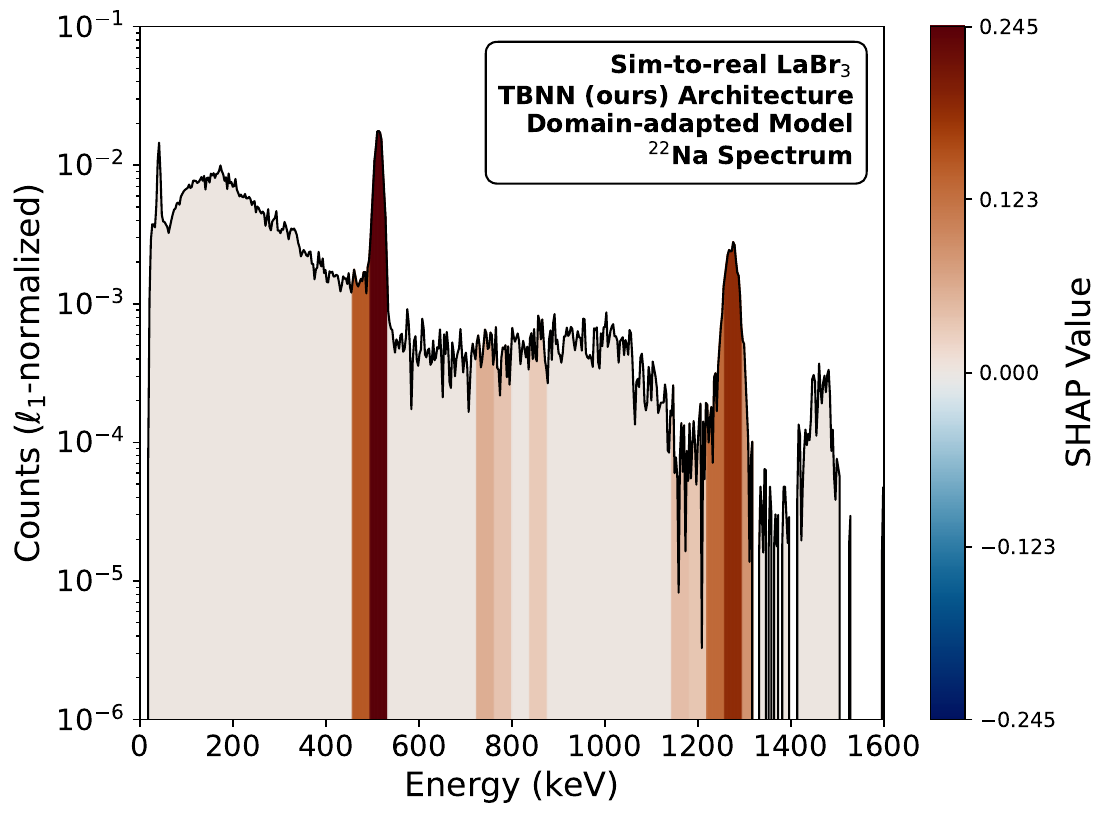}
    \caption{SHAP explanations for an $^{18}$F spectrum (left) and a $^{22}$Na spectrum (right). Both isotopes share a 511 keV annihilation peak, but the model additionally highlights the 1274.5 keV line as salient for $^{22}$Na.}
    \label{fig:XAI_F18_Na22}
\end{figure}

\end{sloppypar}
\end{document}